\documentclass[conference]{IEEEtran}
\IEEEoverridecommandlockouts

\usepackage{amsmath,amssymb,amsfonts}
\usepackage{algorithmic}
\usepackage{graphicx}
\usepackage{textcomp}
\usepackage{listings}
\usepackage{color}

\usepackage{fancybox}

\usepackage{hyperref}
\usepackage{booktabs}
\usepackage{multirow}
\usepackage{array}
\usepackage{tikz}
\usetikzlibrary{arrows.meta, positioning}
\usepackage{pgfplots}
\pgfplotsset{compat=1.17}
\usepackage{adjustbox}

\definecolor{mygray}{rgb}{0.5,0.5,0.5}
\definecolor{myblue}{rgb}{0.1,0.1,0.9}
\definecolor{mylightgray}{rgb}{0.95,0.95,0.95}

\usepackage[backend=biber]{biblatex}
\usepackage{xcolor}
\def\BibTeX{{\rm B\kern-.05em{\sc i\kern-.025em b}\kern-.08em
    T\kern-.1667em\lower.7ex\hbox{E}\kern-.125emX}}
\begin{document}

\title{Empirical Evaluation of Open-Source Large Language Models for Retrieval-Augmented Generation in ESG Domain
}

\author{
\IEEEauthorblockN{Motaz Saad}
\IEEEauthorblockA{\textit{University of Salento} \\
motazk.saad@unisalento.it \\
ORCID: 0000-0002-1080-7276}
\and
\IEEEauthorblockN{Anna Borrelli}
\IEEEauthorblockA{\textit{University of Naples Federico II} \\
anna.borrelli3@unina.it}
\and
\IEEEauthorblockN{Ivan Gentile}
\IEEEauthorblockA{\textit{IFAB Foundation} \\
ivan.gentile@ifabfoundation.org}
\and
\IEEEauthorblockN{Kianna Kazemi}
\IEEEauthorblockA{\textit{IFAB Foundation} \\
kianna.kazemi@ifabfoundation.org}
\and
\IEEEauthorblockN{Francesco Piccialli}
\IEEEauthorblockA{\textit{University of Naples Federico II} \\
francesco.piccialli@unina.it}
\and
\IEEEauthorblockN{Antonella Longo}
\IEEEauthorblockA{\textit{University of Salento} \\
antonella.longo@unisalento.it}
}

\maketitle

\begin{abstract}

Environmental, Social, and Governance (ESG) reporting has emerged as a critical component of corporate accountability, with Large Language Models (LLMs) through Retrieval-Augmented Generation (RAG) showing significant potential for automating KPI extraction from reports. However, the performance characteristics of open-source LLMs in ESG domain-specific tasks remain inadequately understood, creating challenges for informed model selection.

This paper addresses the performance characteristics of open-source LLM deployment in ESG contexts through a structured evaluation framework and an ESG-RAG evaluation resource grounded in 498 real-world ESG reports. We conduct rigorous performance evaluation of seven open-source models ranging from 2B to 30B parameters using a corpus of 498 ESG reports from EU-listed companies (2010--2024), from which 100 synthetic QA pairs are drawn for evaluation from a larger 284-pair persona-based synthetic dataset covering Environmental, Social, and Governance information needs. Our evaluation employs persona-driven query generation and established RAGAS metrics including contextual recall, precision, relevance, faithfulness, answer relevancy, and factual correctness.

Our evaluation of glm-4.7-flash (30B MoE), nemotron-3-nano:4b (4B), qwen3:4b-instruct (4B), gemma3:4b (4B), gemma4:e4b (8B), gemma4:e2b (2B), and ministral-3:8b (8B) reveals notable differences in performance characteristics across different model architectures and sizes. Retrieval is strong but not perfect under LLM-based judging (context recall $\approx$0.58--0.61, context precision $\approx$0.78--0.81, context relevance 0.965--0.985, LLM-judged context precision with reference $\approx$0.83--0.88). Generation diverges most on faithfulness (0.607--0.822) and least on answer relevancy (0.760--0.881): glm-4.7-flash leads faithfulness (0.822), qwen3 leads factual correctness (0.449), and ministral-3 leads answer relevancy (0.881). Factual correctness across all models (0.387--0.449) highlights ongoing opportunities for domain-specific fine-tuning. This evaluation provides organizations with data-driven insights for selecting open-source models tailored to their specific ESG reporting requirements.

\end{abstract}


\section{Introduction}

ESG reporting refers to the disclosure of data related to a company's environmental, social, and governance practices. Originally rooted in ethical investing, it has evolved into a key framework for assessing non-financial risks and performance that can influence a company’s long-term sustainability and investor attractiveness \cite{rusu2024sustainability}. ESG reporting demonstrates a company's commitment to sustainability, ethical conduct, and regulatory compliance. Investors, policymakers, and other stakeholders increasingly depend on ESG disclosures to assess corporate performance beyond traditional financial indicators, making the accurate and efficient extraction of ESG Key Performance Indicators (KPIs) a pressing priority. However, ESG data is often complex, heterogeneous, and embedded in diverse, unstructured documents, creating significant challenges in achieving consistency and reliability.


The environmental component of the ESG evaluates how a company affects and is affected by the natural world, focusing on metrics such as carbon emissions (including Scope 1, 2, and 3), energy use, water consumption, waste management, and climate risk strategies. The social dimension centers on the company’s relationships with key stakeholders, including employees, suppliers, and communities. It addresses labor practices, workplace safety, human rights, diversity and inclusion, community involvement, and consumer protection. The governance aspect pertains to the systems and structures guiding corporate behavior, including board composition and diversity, executive compensation, transparency, shareholder rights, and anti-corruption policies. Together, these three pillars form a comprehensive lens through which organizations are evaluated for long-term sustainability and ethical stewardship \cite{orsolin2024esg}.

Another dimension of complexity is that there are different ESG Reporting Standards and Frameworks. Table \ref{tab:ESG_frameworks} summarize some of the most common frameworks. 

\begin{table}[ht]
\centering
\caption{Summary of Major ESG Reporting Frameworks}
\begin{tabular}{|p{1.3cm}|p{5.7cm}|}
\hline
\textbf{Framework} & \textbf{Description} \\
\hline
\textbf{GRI \cite{GRI2021}} & Global Reporting Initiative: Most widely used ESG framework, focused on a company’s overall sustainability impact across environmental, social, and economic dimensions. \\
\hline
\textbf{SASB \cite{SASB2017}} & Sustainability Accounting Standards Board: Provides industry-specific standards that highlight financially material ESG issues relevant to investors. \\
\hline
\textbf{TCFD \cite{TCFD2017}} & Task Force on Climate-related Financial Disclosures: Offers guidance on disclosing climate-related financial risks and opportunities, including governance and scenario analysis. \\
\hline
\textbf{CDP \cite{cdp2025}} & Carbon Disclosure Project: Specializes in environmental disclosures, particularly greenhouse gas emissions, water security, and forest risks. \\
\hline
\textbf{CSRD \cite{EU_CSRD_2022} } & Corporate Sustainability Reporting Directive: European Union regulation requiring standardized ESG disclosures from large companies with third-party assurance and digital tagging. \\
\hline
\textbf{ESRS \cite{ESRS2023}} & European Sustainability Reporting Standards: 12 EFRAG-developed standards (ESRS 1--2 cross-cutting, ESRS E1--E5 environmental, ESRS S1--S4 social, ESRS G1 governance) that operationalise CSRD disclosure requirements under a double-materiality lens. \\
\hline
\end{tabular}
\label{tab:ESG_frameworks}
\end{table}

Artificial Intelligence (AI), especially Large Language Models (LLMs), has emerged as a powerful tool for automating ESG KPI extraction, streamlining data processing, and improving reporting accuracy. Retrieval-Augmented Generation (RAG) techniques further enhance this capability by integrating relevant external data sources into LLM outputs, thereby reducing hallucination and increasing the precision of ESG insights.

Open-source models offer compelling advantages for ESG applications, including deployment flexibility and customization potential. However, the performance characteristics of open-source models in ESG domain-specific tasks remain inadequately understood, creating challenges for informed model selection.

This paper addresses this gap by conducting a comprehensive empirical evaluation of open-source LLMs within RAG frameworks applied to ESG KPI extraction. It provides an assessment framework that evaluates model performance across multiple dimensions including contextual understanding, faithfulness, and factual correctness, enabling organizations to select appropriate open-source solutions tailored to their specific ESG reporting needs.

The paper aims to (1) highlight the growing importance of AI-driven ESG KPI extraction in enhancing sustainability reporting, (2) evaluate the performance characteristics of open-source LLMs in ESG domain-specific RAG applications through systematic benchmarking, (3) analyze the practical implications and deployment considerations of employing open-source models for this purpose, and (4) offer comparative insights to guide organizations toward technically effective open-source solutions for their ESG reporting requirements.

By focusing on both the performance and practical dimensions of open-source AI adoption in ESG analytics, this paper provides valuable guidance for decision-makers seeking to balance technical capabilities with deployment flexibility and optimize resource allocation in sustainability initiatives. The evaluation framework enables organizations to make informed decisions that account for both the quality of ESG information extraction and the practical advantages of open-source solutions, ensuring sustainable and scalable AI deployment strategies.

This work makes two contributions. First, it provides an empirical evaluation of five open-source LLMs for ESG-focused RAG under a controlled retrieval setup. Second, it introduces an ESG-RAG evaluation resource (available at \url{https://github.com/motazsaad/ESG-RAG-benchmark-ragas}) consisting of a processed corpus of 498 ESG reports, 284 synthetic QA pairs grounded in source documents with persona-based coverage across ESG pillars, and accompanying evaluation outputs and scripts to support reproducible comparison.

\section{State of the Art}

\begin{figure}[ht]
\centering
\begin{tikzpicture}[
  node distance=1.1cm and 0.8cm,
  box/.style={rectangle, draw=blue!60, fill=blue!10, thick, minimum height=0.7cm, minimum width=2.0cm, text centered},
  data/.style={rectangle, draw=orange!70, fill=orange!10, thick, minimum height=0.5cm, minimum width=2cm, text centered},
  arrow/.style={thick, ->, >=Stealth}
  ]

\node[box] (user) {User Query};
\node[box, right=of user] (retriever) {Retriever};
\node[data, above=of retriever] (docstore) {Document Store};

\node[box, below=of retriever] (generator) {LLM Generator};
\node[box, right=of generator] (output) {Final Answer};

\draw[arrow] (user) -- (retriever);
\draw[arrow] (retriever) -- (docstore);
\draw[arrow] (docstore) -- (retriever) node[midway, above ] {\tiny Top-K Docs};
\draw[arrow] (retriever) -- (generator);
\draw[arrow] (generator) -- (output);

\end{tikzpicture}
\caption{RAG pipeline}
\label{fig:RAG}
\end{figure}
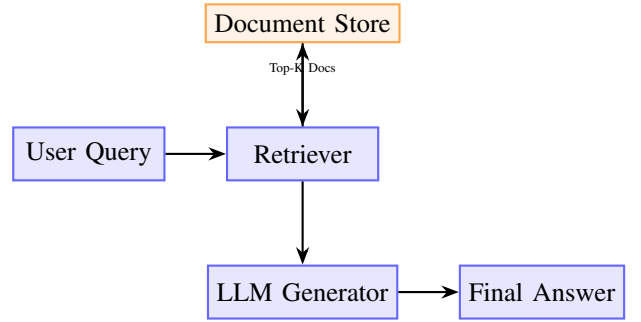

Retrieval-Augmented Generation (RAG) is a cutting-edge architecture in NLP that fuses document retrieval with language generation, addressing a key limitation of large language models (LLMs): their inability to access external, up-to-date knowledge at inference time. Originally proposed by Facebook AI Research, RAG systems leverage real-time access to external corpora, enabling models to generate more factually accurate, context-aware answers \cite{lewis2020retrieval}. Figure \ref{fig:RAG} shows the key components of a typical RAG system: 

\begin{itemize}
    \item Retriever: Selects relevant documents. The retrieval quality directly impacts generation accuracy. 

    \item Generator: A pretrained LLM that synthesizes an output based on the retrieved documents plus the query.
\end{itemize}

The Mechanism of Retrieval-Augmented Generation (RAG) is as follows: 

\begin{enumerate}
    \item \textbf{User Query:} Let the input be a natural language query \( q \).
    
    \item \textbf{Retriever:} A dense retriever encodes the query into a vector and retrieves the top \( k \) documents from an external corpus \( D \). This can be formalized as:
    \[
    \{d_i\}_{i=1}^k = \text{Retriever}(q, D)
    \]
    where each \( d_i \) is a document retrieved based on semantic similarity.
    
    \item \textbf{Generator:} For each retrieved document \( d_i \), the generator model produces a response conditioned on the pair \( (q, d_i) \):
    \[
    y_i = \text{Generator}(q, d_i)
    \]
\end{enumerate}

This architecture enables dynamic knowledge integration without retraining the generator, as updating the external data is sufficient to refresh the model's knowledge.

RAG has evolved beyond the original retrieve-then-generate paradigm. Gao et al.\ \cite{gao2023rag} categorize its development into three generations—Naive, Advanced, and Modular RAG—showing how pre-retrieval query rewriting, post-retrieval re-ranking, and adaptive generation progressively improve factual grounding. These architectural advances make RAG the preferred approach for knowledge-intensive tasks requiring access to specialized corpora such as ESG reports.

\subsection{RAG Evaluation Frameworks}

Systematic RAG evaluation has been formalized through dedicated frameworks. Es et al.\ propose RAGAS \cite{ragas2024}, a reference-free evaluation library measuring faithfulness, answer relevancy, context precision, and context recall using LLM-as-judge scoring---the framework adopted in this study. Saad-Falcon et al.\ introduce ARES \cite{ares2024}, which fine-tunes lightweight LM judges on synthetic training data for automated pipeline assessment. In addition to research-oriented frameworks such as RAGAS and ARES, practitioner-oriented tooling such as DeepEval supports end-to-end and component-level RAG evaluation with metrics for answer relevancy, faithfulness, contextual relevancy, precision, and recall \cite{deepeval_docs}. More broadly, the use of LLMs as evaluators has been studied in general NLG and LLM benchmarking settings through frameworks such as G-Eval \cite{liu2023geval}, JudgeLM \cite{zhu2025judgelm}, and Prometheus \cite{kim2024prometheus}, while recent survey work highlights both the practical value of LLM-as-a-judge systems and their reliability challenges \cite{gu2024llmjudge}. Lima et al.\ \cite{lima2025rag} provide a taxonomic framework for synthetic benchmark generation in domain-specific RAG, demonstrating that dataset diversity and generation strategy substantially affect evaluation validity---a finding that directly motivates the persona-driven, quality-filtered protocol used in our evaluation dataset construction.

\subsection{LLMs and RAG for ESG Applications}

The application of LLMs to ESG analysis has attracted growing research attention across extraction, generation, and benchmarking tasks. ESGReveal \cite{esgreveal2025} applies RAG with GPT-4 to extract structured ESG data from 166 Hong Kong company reports, demonstrating strong retrieval-augmented extraction performance but evaluating only a single proprietary model. Climate Finance Bench \cite{climatefinbench2025} benchmarks RAG pipelines over corporate climate disclosure PDFs with expert-annotated QA pairs, finding that even the strongest proprietary system achieves only 62\% accuracy—highlighting task difficulty and leaving open-source models entirely unevaluated.

On the open-source side, SusGen-GPT \cite{susgen2025} fine-tunes 7--8B open-source LLMs on a multi-task ESG and financial corpus, showing that small open-source models can approach proprietary performance on sustainability report generation, but without evaluating them within a RAG pipeline. Birti et al.\ \cite{birti2025esg} optimize Llama and Gemma models for ESG activity detection and EU taxonomy classification, confirming competitive open-source performance on classification tasks, again without a RAG evaluation setting.

Benchmarks like ESGenius \cite{ESGenius25} and MMESGBench \cite{MMESGBench25} demonstrate that RAG significantly boosts LLM performance on ESG tasks, especially for smaller models, while multimodal approaches outperform text-only systems on visually grounded disclosures. Fine-tuning datasets such as ESG-CID \cite{ESGCID25} enable domain-specific retrieval optimization. Practical tools like ESG-Consultant \cite{ESGConsultant25} show promise in compliance support, while ESGLens \cite{esglens2026} extends ESG-focused RAG toward interactive report analysis with source traceability and score prediction. Studies in low-resource languages—such as Turkish ESG report analysis—reveal that traditional IR methods like BM25 can outperform transformer-based retrievers when paired with proprietary generators \cite{RAGTR}.

\subsection{Research Gap}

Despite these advances, there remains limited work comparing multiple open-source LLMs as generators within a shared ESG RAG pipeline under controlled retrieval, prompting, and evaluation conditions. Existing work either benchmarks ESG knowledge more broadly with RAG as one condition rather than as a controlled cross-model pipeline study \cite{ESGenius25,MMESGBench25}, introduces ESG QA benchmarks or domain resources without a shared open-source generator comparison over one pipeline configuration \cite{ESGCID25}, or presents ESG-focused RAG systems without systematic cross-family generator evaluation \cite{esgreveal2025,climatefinbench2025,esglens2026}. This gap makes it difficult for organizations to make evidence-based decisions about open-source model selection for ESG applications under practical local-deployment constraints.

\subsection{ESG Reports: Structure and Data Extraction Challenges}

ESG reports follow established frameworks such as GRI, SASB, TCFD, and ESRS \cite{cdp2025,GRI2021,EU_CSRD_2022} to ensure comparability and transparency while focusing on material topics most relevant to stakeholders.

\textbf{Key Structural Elements:} ESG reports typically include executive summaries with CEO messages; company profiles and ESG strategy alignment; governance structures and risk management frameworks; materiality assessments identifying priority topics; environmental performance data (emissions, resource use, climate targets); social performance metrics (workforce diversity, human rights, community impact); governance disclosures (board composition, executive compensation tied to ESG); quantitative KPIs with benchmarking; framework alignment tables; and third-party assurance statements.

\textbf{Data Presentation Formats:} ESG reports employ diverse layouts including structured tables for KPI comparisons and framework mappings, infographics and visual narratives for stakeholder engagement, interactive dashboards in digital formats, materiality matrices plotting stakeholder concerns against business impact, charts and graphs for trend visualization, and geographic maps showing operational impacts.

\textbf{Data Extraction Challenges:} Extracting ESG KPIs from these varied formats presents significant automated processing challenges. Key difficulties include unstructured narrative text with non-standardized phrasing, multi-modal content combining text and visuals that resist traditional parsing methods, inconsistent table formatting across companies and sectors, visual data encoding through charts and color-coding, and variable language expression where identical metrics may be described differently (e.g., "CO2 footprint reduced by 15\%" versus "Scope 1 emissions fell by 15\%").

A typical extraction challenge occurs when emission values appear within pie charts without clear legends, while accompanying narratives provide only vague statements like "value-chain emissions saw a significant drop," with actual figures relegated to footnotes. This complexity necessitates advanced techniques combining visual parsing, natural language understanding, and contextual inference for effective ESG KPI retrieval.

\section{Methodology}

\subsection{Evaluation framework (RAGAS)}

We evaluate with RAGAS \cite{ragas2024}, an open-source library designed specifically for RAG assessment. RAGAS provides metric implementations for: (i) generation quality—faithfulness (does the answer stay grounded in retrieved context?) and answer relevancy (does it answer the user’s question?); and (ii) retrieval quality—context precision (signal-to-noise in retrieved chunks) and context recall (were all relevant chunks retrieved). We use the library’s evaluate() API to compute scores over our evaluation set (details below). RAGAS uses an LLM-judge to score most metrics (reference-free), while context recall relies on ground-truth annotations or synthesized gold answers to estimate true positives and false negatives. The evaluation metrics that we use in our evaluation include: 

\begin{itemize}
    \item Faithfulness: penalizes hallucinations by checking whether answer claims are supported by the retrieved passages.
    \item Answer relevancy: measures how directly and completely the answer addresses the query.
    \item Context precision: fraction of retrieved content that is actually useful (high precision leads to less distractor text).
    \item Context recall: fraction of all relevant content that was retrieved (high recall leads to fewer misses).
\end{itemize}

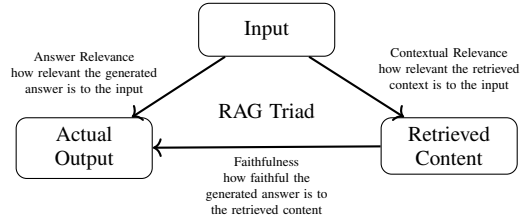
\begin{figure}
\centering
\begin{tikzpicture}[
    node distance = 0.8cm and 0.6cm,
    scale=0.8,
    box/.style = {rectangle, draw, rounded corners, minimum width=1.8cm, minimum height=0.7cm, align=center, font=\footnotesize},
    arrow/.style = {->, thick}
]

\node[box] (input) {Input};
\node[box, below left=of input] (actual) {Actual\\Output};
\node[box, below right=of input] (retrieved) {Retrieved\\Content};

\node[below=-1.8cm of actual, font=\tiny, text width=2cm, align=center] 
    {Answer Relevance\\
    how relevant the generated answer is to the input};

\node[below=-1.8cm of retrieved, font=\tiny, text width=2cm, align=center] 
    {Contextual Relevance\\
    how relevant the retrieved context is to the input};

\node[below=0.5cm of input, font=\footnotesize, text width=1.8cm, align=center] 
    {RAG Triad};

\node[below=1.2cm of input, font=\tiny, text width=2.2cm, align=center] 
    {Faithfulness\\
    how faithful the generated answer is to the retrieved content};

\draw[arrow] (input) -- (actual);
\draw[arrow] (input) -- (retrieved);
\draw[arrow] (retrieved) -- (actual);

\end{tikzpicture}
\caption{RAG Triad}
\label{fig:RAG_Triad}
\end{figure}

Together, these metrics expose common failure modes: low precision (over-stuffed context), low recall (missed evidence), low faithfulness (hallucination), and low answer relevancy (off-topic or incomplete answers). The RAG triad as shown in Figure \ref{fig:RAG_Triad} is composed of three RAG evaluation metrics: answer relevancy, faithfulness, and contextual relevancy. Each metric corresponds to a specific component of the RAG pipeline, making the triad a useful diagnostic tool for identifying which component limits overall performance. Beyond these four core metrics, our evaluation further incorporates LLM Context Precision With Reference, Context Relevance, and Factual Correctness; the complete set of seven metrics is detailed in Section~\ref{sec:evaluation_metrics}.

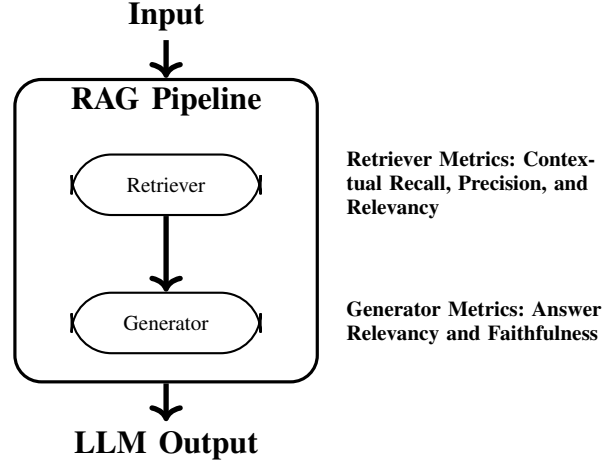
\begin{figure}
\centering
\begin{tikzpicture}[
    node distance = 1.2cm and 2cm,
    scale=0.85,
    pipeline/.style = {rectangle, draw, very thick, rounded corners=10pt, minimum width=4cm, minimum height=4cm, align=center},
    component/.style = {rectangle, draw, thick, rounded corners=15pt, minimum width=2.5cm, minimum height=0.8cm, align=center, font=\footnotesize},
    label/.style = {font=\large\bfseries},
    arrow/.style = {->, very thick, line width=2pt}
]

\node[label] (input) {Input};

\node[pipeline, below=0.5cm of input] (pipeline) {};
\node[label, anchor=north] at (pipeline.north) {\vspace{0.3cm}RAG Pipeline};

\node[component, below=1cm of pipeline.north] (retriever) {Retriever};
\node[component, below=1cm of retriever] (generator) {Generator};

\node[label, below=0.5cm of pipeline] (output) {LLM Output};

\draw[arrow] (input) -- (pipeline.north);
\draw[arrow] (retriever) -- (generator);
\draw[arrow] (pipeline.south) -- (output);

\node[right=1cm of retriever, anchor=west, text width=3.5cm, font=\footnotesize] (ret_metrics) {
    \textbf{Retriever Metrics: Contextual Recall, Precision, and Relevancy}\\
};

\node[right=1cm of generator, anchor=west, text width=3.5cm, font=\footnotesize] (gen_metrics) {
    \textbf{Generator Metrics: Answer Relevancy and Faithfulness }\\
};

\end{tikzpicture}
\caption{RAG Evaluation}
\label{fig:RAG_Evaluation}
\end{figure}

Figure~\ref{fig:RAG_Evaluation} illustrates that RAG evaluation requires measuring both components separately. Metrics must ensure the retriever finds the right information and that the generator produces accurate, relevant answers based on that retrieved context. This aligns with the synthetic data generation process described in Section~\ref{sec:synth_validity}, where question-answer pairs are created along with the relevant source chunks to evaluate both retrieval and generation performance.

\subsection{Dataset Selection and Preparation}
\label{sec:dataset_prep}

Our evaluation utilizes a dataset of 498 ESG reports collected from publicly available corporate websites, spanning 15 years (2010-2024) of sustainability reporting from EU-listed companies. This corpus provides coverage across different years and document sizes, enabling robust evaluation of model performance across diverse ESG reporting contexts.

\subsubsection{Selection Strategy}

The document selection process followed a systematic approach intended to provide broad coverage across years and document sizes:

\begin{itemize}
    \item \textbf{Temporal Coverage}: Documents were sampled to ensure balanced representation across all 15 years, with 6-7 documents selected per year depending on corpus availability.
    \item \textbf{Size Diversity}: Documents were selected from different size quartiles to ensure variety in document complexity, with selected documents ranging from 11,137 to 1,482,863 bytes.
    \item \textbf{Random Selection}: Within each stratum (year and size category), random selection was used to reduce manual selection bias.
\end{itemize}

The resulting dataset provides broad coverage of ESG reporting evolution while maintaining manageable computational requirements for multiple model evaluations.

\subsubsection{Dataset Characteristics}

The selected 498-document dataset exhibits the following characteristics:

\begin{itemize}
    \item \textbf{Temporal Distribution}: Balanced coverage across 2010-2024, enabling evaluation of model performance on ESG reporting evolution.
    \item \textbf{Size Distribution}: Mean size of 282,119 bytes with median of 230,348 bytes, providing variety in document complexity.
    \item \textbf{Content Diversity}: Covers environmental, social, and governance topics across different industries and reporting frameworks.
\end{itemize}

\subsubsection{PDF-to-Markdown Conversion}

The source corpus consists of PDF files. Prior to indexing, each document was converted to plain-text Markdown using a custom pipeline built on PyMuPDF (\texttt{fitz}). The pipeline proceeds as follows:

\begin{enumerate}
    \item \textbf{Block extraction}: Each page is processed via PyMuPDF's \texttt{get\_text("blocks")} API, which returns text blocks with bounding-box coordinates, preserving the spatial layout of the original PDF.

    \item \textbf{Heading detection}: Blocks are classified as headings using two heuristics: (a) text length under 50 characters with at most 8 words, or (b) all-uppercase text under 100 characters. Heading level is assigned by length and casing---short all-caps blocks become \texttt{\#~H1}, blocks under 50 characters become \texttt{\#\#~H2}, and remaining heading-classified blocks become \texttt{\#\#\#~H3}.

    \item \textbf{Text cleaning}: Body text undergoes whitespace normalization (collapsing multiple spaces), insertion of spaces at lowercase-to-uppercase and digit-to-letter transitions to repair common PDF extraction artefacts, and removal of isolated page-number lines.

    \item \textbf{Page markers}: A separator is appended after each page to preserve page-boundary information for downstream chunking.
\end{enumerate}

This process preserves the logical document structure required for accurate chunk-level retrieval while producing clean plain text suitable for embedding. Tables and figures embedded purely as images are not captured by this text-extraction approach and therefore fall outside the scope of this evaluation.

\subsubsection{Dataset Availability}

The complete ESG-RAG evaluation resource is publicly available at \url{https://github.com/motazsaad/ESG-RAG-benchmark-ragas}. The repository includes all 498 ESG reports in processed Markdown format, the generated query-answer pairs in CSV and JSON formats, complete RAGAS evaluation scores for all seven models, document selection metadata, and evaluation scripts for reproducing the experimental setup.

\subsection{Synthetic Data Generation for RAG Evaluation}

\begin{figure}[ht]
\centering
\resizebox{\columnwidth}{!}{%
\begin{tikzpicture}[
  sbox/.style={rectangle, draw=blue!60, fill=blue!8, thick,
               minimum width=2.2cm, minimum height=0.75cm,
               align=center, font=\footnotesize},
  obox/.style={rectangle, draw=green!60!black, fill=green!10, thick,
               minimum width=2.2cm, minimum height=0.75cm,
               align=center, font=\footnotesize},
  arrow/.style={thick,->,>=Stealth}
]
\node[sbox] (s1)  at (0,0)       {\textbf{1.} Load\\Data};
\node[sbox] (s2)  at (3.1,0)     {\textbf{2.} Chunk\\Data};
\node[sbox] (s3)  at (3.1,-1.7)  {\textbf{3.} Generate\\Questions};
\node[sbox] (s4)  at (0,-1.7)    {\textbf{4.} Generate\\Answers};
\node[sbox] (s5)  at (0,-3.4)    {\textbf{5.} Quality\\Filter};
\node[obox] (out) at (3.1,-3.4)  {Evaluation\\Dataset};

\draw[arrow] (s1)  -- (s2);
\draw[arrow] (s2)  -- (s3);
\draw[arrow] (s3)  -- (s4);
\draw[arrow] (s4)  -- (s5);
\draw[arrow] (s5)  -- (out);
\end{tikzpicture}%
}
\caption{Five-step synthetic dataset generation protocol for RAG evaluation~\cite{ragas2024,bedrock_synth_rag2024}. Data flows in a snake pattern: documents are loaded and chunked (Steps~1--2); questions and reference answers are derived \emph{directly} from each sampled chunk (Steps~3--4); a length-based quality filter removes degenerate pairs (Step~5).}
\label{fig:synth_pipeline}
\end{figure}
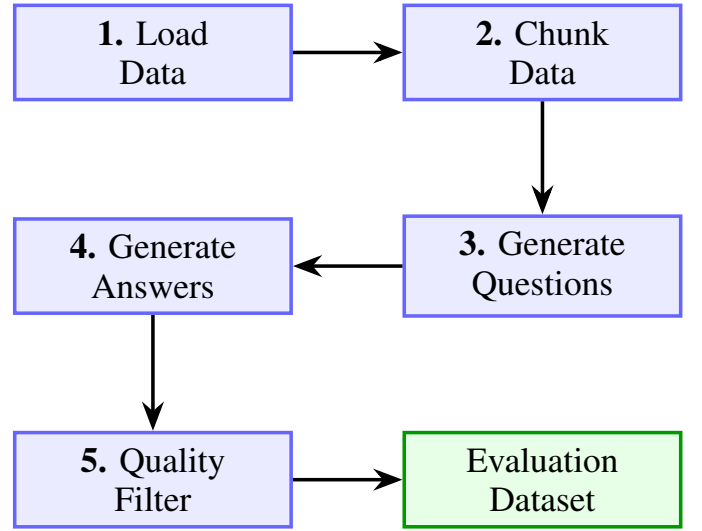

Evaluating a RAG pipeline requires a labeled dataset containing (i) queries that reflect real user information needs and (ii) reference answers grounded in the source documents. Since no such labeled benchmark existed for our ESG corpus described in Section~\ref{sec:dataset_prep}, we constructed one using a custom synthetic generation protocol inspired by prior RAG evaluation workflows~\cite{ragas2024,bedrock_synth_rag2024}, illustrated in Figure~\ref{fig:synth_pipeline}. We used \texttt{qwen3:4b-instruct} as the generator; this model also appears in the evaluated set, which introduces a self-enhancement risk that we quantify in Section~\ref{sec:limitations}.

\subsubsection{Validity of the Synthetic Evaluation Protocol}
\label{sec:synth_validity}

A legitimate concern with synthetic evaluation datasets is whether machine-generated QA pairs can accurately reflect real-world RAG performance. In our protocol, every question and reference answer is derived solely from the real 498-report ESG corpus, and each retained QA pair records its originating document source. This keeps the benchmark grounded in actual disclosures and preserves document-level provenance for later audit and retrieval analysis.

The protocol also applies simple but explicit controls to reduce degenerate samples. Source chunks shorter than 50 characters are discarded before generation, and generated QA pairs are retained only when both the question and the reference answer exceed 10 characters. Combined with persona rotation across the Environmental, Social, and Governance pillars, these design choices are intended to produce a benchmark that captures persona-guided ESG information needs while maintaining broad topical coverage.

At the same time, the benchmark should be understood as a custom RAGAS-inspired synthetic resource rather than a direct implementation of the full RAGAS testset-generation pipeline. The dataset was generated with \texttt{qwen3:4b-instruct}; one of the evaluated models shares this generator family, and the same model family also serves as the RAGAS judge. We therefore treat the benchmark as a practical comparative instrument within the scope of this study and explicitly acknowledge the resulting self-enhancement risk in Section~\ref{sec:limitations}.

\subsubsection{Generation Protocol}

The pipeline proceeds in five steps:

\begin{enumerate}
    \item \textbf{Load Data}: Each ESG document in the 498-report corpus is loaded from the document store.

    \item \textbf{Chunk Data}: Documents are split into fixed-size chunks of 1000 characters with a 200-character overlap (step size 800 characters). Chunks shorter than 50 characters after stripping are discarded. This produces a knowledge graph of 258{,}590 chunks across the 498 reports. From this pool we randomly sample 300 chunks (seed 42) for QA generation; personas (Section~\ref{sec:personas}) are rotated cyclically across the sample so each ESG pillar receives roughly one third of the questions.

    \item \textbf{Generate Questions}: An LLM analyzes each sampled chunk and generates a natural-language question that is (a) answerable solely from the chunk's content and (b) representative of a genuine ESG information need. Questions are constrained to avoid compound queries that conflate multiple retrieval targets.

    \item \textbf{Generate Answers}: Using the same chunk as the sole context, the LLM generates a concise reference answer. This serves as the ground truth against which the RAG system's output is compared by RAGAS metrics.

    \item \textbf{Quality Filter}: Degenerate outputs are removed using minimum length thresholds. Source chunks shorter than 50 characters are discarded before generation. Generated QA pairs are retained only if both the question and the reference answer each exceed 10 characters, filtering out empty or truncated outputs.
\end{enumerate}

\subsubsection{Persona-Driven Query Specialization}
\label{sec:personas}

To ensure domain coverage across all three ESG pillars, we layer the RAGAS persona framework~\cite{ragas2024} on top of the generation protocol. Each persona directs the question generation LLM toward a specific stakeholder perspective:

\begin{lstlisting}[language=Python, caption=ESG Persona Configuration]
from ragas.testset.persona import Persona
persona_environmental_kpi_extractor = Persona(
    name="Environmental KPI Extractor",
    role_description="Focuses on extracting key environmental performance
    indicators (KPIs) from documents, such as carbon emissions, water
    consumption, energy usage, waste generation, and renewable energy adoption.")
persona_social_kpi_extractor = Persona(
    name="Social KPI Extractor",
    role_description="Focuses on extracting key social performance indicators
    (KPIs) from documents, such as employee diversity, labor practices,
    community engagement, human rights, and health and safety metrics.")
persona_governance_kpi_extractor = Persona(
    name="Governance KPI Extractor",
    role_description="Focuses on extracting key governance performance
    indicators (KPIs) from documents, such as board diversity, executive
    compensation, anti-corruption policies, shareholder rights, and data
    privacy measures.")
\end{lstlisting}

The combination of the structured generation protocol, length-based filtering, and persona-driven specialization produces an evaluation dataset that (i) is fully grounded in real ESG source documents, (ii) covers all three ESG pillars, (iii) contains traceable evidence links for retrieval validation, and (iv) captures persona-guided ESG information needs---making it a rigorous benchmark for RAG evaluation in the sustainability reporting domain.

After length filtering, the protocol yields 284 QA pairs (Environmental: 97, Social: 94, Governance: 93) drawn from 181 unique source documents out of the 498 in the corpus.

\subsection{Open-Source Model Selection}

Our evaluation focuses on open-source models ranging from 2B to 30B parameters. This band spans the five evaluated checkpoints (plus the Gemma 4 models in the evaluation framework) and remains feasible in our local deployment setting, allowing direct comparison across compact 2B--4B-class models, 8B-class models, and 30B MoE models under the same 48 GB GPU budget.

\subsubsection{Selection Criteria and Rationale}

The model selection strategy employed a multi-criteria approach to provide practical coverage of several widely used open-source model families under a common local-deployment setup:

\begin{itemize}
    \item \textbf{Industry Adoption}: Models selected based on widespread adoption in enterprise and research settings, as evidenced by leaderboard rankings and community usage statistics
    \item \textbf{Architectural Diversity}: Models from different families (GLM, Qwen, Gemma, Ministral, Nemotron) to evaluate architectural impacts and training methodology differences
    \item \textbf{Geographic Representation}: Models from major AI research hubs (US, Europe, China) to ensure global perspective
    \item \textbf{Size Range}: 2B-30B parameter range covering edge (gemma4:e2b) to workstation-class MoE models (glm-4.7-flash), while remaining feasible on the available hardware
    \item \textbf{Platform Accessibility}: All models accessible through Ollama platform for consistent evaluation environment and reproducibility
\end{itemize}

\subsubsection{Evaluated Models}

The evaluation included seven open-source models selected based on their performance in industry benchmarks and research literature:

\begin{itemize}
    \item \textbf{glm-4.7-flash:q4\_K\_M}: 30 billion parameter Mixture-of-Experts model from Zhipu AI (3B active), demonstrating strong performance in Chinese language tasks and ranking among top models in the Open LLM Leaderboard \cite{openllmleaderboard}. The q4\_K\_M quantization format has been reported to reduce memory footprint substantially while preserving most of the original model performance \cite{frantar2023qlora}.
    
    \item \textbf{qwen3:4b-instruct}: 4 billion parameter model from Alibaba, representing the latest generation of the Qwen series which has shown exceptional performance in multilingual tasks \cite{qwen3tech}. Selected for its balanced performance across English and Chinese ESG reporting contexts.
    
    \item \textbf{gemma3:4b}: 4 billion parameter model from Google, representing the latest generation architecture with significant improvements over previous Gemma models \cite{team2024gemma}. Chosen for its strong performance in reasoning tasks and Google's extensive research backing in language model development.
    
    \item \textbf{gemma4:e4b}: 8 billion parameter model from Google DeepMind (4.5B effective), designed for frontier intelligence locally with configurable thinking modes \cite{team2024gemma}. Selected to represent the latest generation of Gemma architecture with enhanced reasoning and agentic capabilities.
    
    \item \textbf{gemma4:e2b}: 2 billion parameter model from Google DeepMind (2.3B effective), optimized for edge device deployment \cite{team2024gemma}. Selected to evaluate efficiency-focused deployment scenarios.
    
    \item \textbf{ministral-3:8b}: 8 billion parameter model from Mistral AI, the latest generation dense model with state-of-the-art performance in edge deployment scenarios \cite{ministral3}. Selected as the European representative and for its strong performance in multilingual contexts.
    
    \item \textbf{nemotron-3-nano:4b}: 4 billion parameter model from NVIDIA, featuring a hybrid Mamba-Transformer MoE architecture optimized for enterprise agentic applications \cite{nemotron3nano}. Selected for its enterprise-grade design and NVIDIA's extensive ecosystem support.
\end{itemize}

\subsubsection{Exclusion Criteria}

Several popular models were deliberately excluded to preserve a controlled and computationally feasible comparison. The study design intentionally evaluates representative checkpoints per model family within the practical 2B--30B local-deployment range, under a fixed hardware budget and a single shared pipeline configuration:

\begin{itemize}
    \item \textbf{Llama Series}: Excluded because the study evaluates one representative checkpoint per family rather than multiple checkpoints from the same lineage; broader family coverage, including Llama, is planned as future work \cite{meta2023llama}
    \item \textbf{Larger Models (>30B)}: Models exceeding 30B total parameters excluded to keep the comparison feasible under the available local hardware budget and to avoid runtime disparities dominating the experimental design
    \item \textbf{Specialized Models}: Domain-specific models (e.g., medical, legal) excluded to keep the study focused on general-purpose generators for ESG RAG rather than task-specific adaptation
    \item \textbf{Older Generations}: Models predating 2023 excluded to keep the comparison centered on recent checkpoints available in the current deployment stack
\end{itemize}

This selection strategy ensures broad coverage of current open-source LLM capabilities while maintaining experimental feasibility and reproducibility standards essential for academic research \cite{brown2020language}.

\subsection{Experimental Setup}

\subsubsection{Hardware Configuration}

All experiments were conducted on a system with the following hardware specifications to ensure reproducibility and provide context for computational requirements:

\begin{itemize}
    \item \textbf{CPU}: 64-core Intel QEMU Virtual CPU (x86\_64 architecture)
    \item \textbf{Memory}: 125 GB RAM with 118 GB available memory
    \item \textbf{GPU}: NVIDIA RTX 6000 Ada Generation with 48 GB VRAM
    \item \textbf{GPU Driver}: NVIDIA-SMI 580.126.09 with CUDA 13.0 support
    \item \textbf{Virtualization}: KVM full virtualization environment
\end{itemize}

The GPU configuration provided substantial computational resources for model inference, with the RTX 6000 Ada Generation offering 48 GB of VRAM sufficient for loading and evaluating multiple models up to 30B parameters simultaneously. End-to-end runtime was approximately 21 hours and 30 minutes for the original five models: about 9 hours 34 minutes for Step 2 (RAG generation across five models on 100 QA pairs over the 498-document corpus) and about 11 hours 56 minutes for Step 3 (RAGAS LLM-judged scoring of seven metrics over the same five model$\times$100 QA matrix). The two additional Gemma 4 models were evaluated in a separate batch under the same pipeline configuration, extending the total runtime accordingly.

\subsubsection{Evaluation Subset}

To bound computational cost, RAG evaluation was run on a deterministic 100-item prefix of the 284-pair synthetic testset. The remaining 184 QA pairs were not used in the experiments reported here and are reserved for future sensitivity analyses. Across the seven evaluated models this yields 700 model--query evaluations.

\subsubsection{Prompt Format}
\label{sec:prompt_format}

All five evaluated models receive identical single-turn prompting. The system message instructs the generator to ground its answer in the retrieved context rather than draw on parametric knowledge: \emph{``You are a helpful assistant that answers questions based on given documents only.''} The human-turn template follows the Python f-string

\begin{lstlisting}[language=Python, basicstyle=\ttfamily\footnotesize, frame=single, breaklines=true]
"question: {query}\n\nDocuments: {relevant_doc}"
\end{lstlisting}

where \texttt{relevant\_doc} is the single top-1 retrieved chunk wrapped in a one-element Python list. No few-shot exemplars, chain-of-thought scaffolding, or persona conditioning are applied at inference time. Because the system message, user template, retrieved context, and retriever are held constant, the main experimental variable is the generator model itself. However, decoding parameters were not explicitly normalized across models in Step 2, so generation behavior may also reflect model-specific default sampling settings exposed through Ollama.

\subsubsection{Embedding Standardization}

To ensure consistent evaluation across all models, we standardized the embedding component using qwen3-embedding:4b for both dataset generation and RAG evaluation. This approach eliminates embedding model variability, allowing us to focus evaluation on generation model performance differences.

\subsubsection{Retrieval Configuration}
\label{sec:retrieval_config}

Retrieval is configured at top-$k$=1: for each query the system computes cosine similarity against all 258{,}590 corpus-chunk embeddings and returns only the single highest-scoring chunk as the entire context window for the generator. There is no reranker, no MMR diversification, no similarity threshold, and no top-$k$$>$1 fusion. This deliberately minimal-context configuration isolates the generator's behaviour on a fixed evidence window and eliminates cross-chunk fusion as a confound when comparing model families, at the cost of imposing a retrieval ceiling that we quantify in Section~\ref{sec:results_retrieval} and acknowledge in Section~\ref{sec:limitations}. Larger top-$k$, hybrid retrievers (BM25\,+\,dense), and reranker variants are left as future work.

\subsubsection{Evaluation Metrics}
\label{sec:evaluation_metrics}

Our evaluation employs the RAGAS framework with seven key metrics:

\begin{itemize}
    \item \textbf{LLM Context Precision With Reference}: LLM-judged variant of context precision. Given the reference answer and the retrieved contexts, the judge decides which retrieved chunks contain information useful for deriving the reference, then averages the chunk-level verdicts into a precision score.
    \item \textbf{Context Precision}: Proportion of retrieved context that is actually relevant.
    \item \textbf{Context Recall}: Ability to retrieve all relevant context from the document corpus.
    \item \textbf{Context Relevance}: Relevance of retrieved context to the query.
    \item \textbf{Faithfulness}: Consistency of generated answers with retrieved context.
    \item \textbf{Answer Relevancy}: Relevance of generated answers to the original query.
    \item \textbf{Factual Correctness}: F1-style accuracy of information provided in generated answers, scored against the reference.
\end{itemize}

\subsubsection{Evaluation Process}

The evaluation process followed these steps:

\begin{enumerate}
    \item \textbf{Dataset Generation}: Use persona-driven approach to generate evaluation queries from the 498-document dataset.
    \item \textbf{RAG Pipeline Setup}: Configure each model with standardized embedding and retrieval components.
    \item \textbf{Query Execution}: Process evaluation queries through each RAG system.
    \item \textbf{Metrics Computation}: Calculate RAGAS scores for each model-query combination.
    \item \textbf{Performance Analysis}: Aggregate and analyze results across models and metrics.
\end{enumerate}

\subsubsection{Implementation Details}

Table~\ref{tab:impl_details} summarises all hyperparameters and configuration choices used in both the dataset generation and RAG evaluation pipelines, providing the information needed to reproduce the experimental setup.

\begin{table}[h]
\centering
\caption{Implementation hyperparameters for dataset generation and RAG evaluation.}
\label{tab:impl_details}
\begin{tabular}{p{0.31\columnwidth} p{0.61\columnwidth}}
\toprule
\textbf{Parameter} & \textbf{Value} \\
\midrule
\multicolumn{2}{l}{\textit{Dataset Generation}} \\
Generator LLM          & qwen3:4b-instruct \\
Generator context window & 12{,}000 tokens \\
Embedding model        & qwen3-embedding:4b \\
Chunk size             & 1{,}000 characters \\
Chunk overlap          & 200 characters \\
Min.\ chunk length     & 50 characters \\
Min.\ QA pair length   & 10 characters (question and answer) \\
\midrule
\multicolumn{2}{l}{\textit{RAG Evaluation (Step 2: retrieval + generation)}} \\
Embedding model        & qwen3-embedding:4b \\
Retrieval method       & Cosine similarity \\
Top-k                  & 1 \\
Generator temperature  & Not overridden in Step 2; model-specific Ollama defaults therefore apply unless defined otherwise by the served checkpoint \\
System prompt          & ``You are a helpful assistant that answers questions based on given documents only.'' \\
User prompt template   & \texttt{"question: \{query\}\textbackslash n\textbackslash nDocuments: \{relevant\_doc\}"} (see Section~\ref{sec:prompt_format}) \\
\midrule
\multicolumn{2}{l}{\textit{RAGAS Scoring (Step 3: LLM-as-judge)}} \\
Judge LLM              & qwen3:4b-instruct \\
Judge embedding        & qwen3-embedding:4b \\
Judge temperature      & Controlled by the RAGAS LLM wrapper for single-completion scoring (near-deterministic default behavior) \\
Max workers            & 1 (serial execution) \\
Per-call timeout       & 3600 s \\
\bottomrule
\end{tabular}
\end{table}

\section{Results: Open-Source LLM Performance Analysis in RAG Application for ESG Domain}

This section presents the evaluation results of seven open-source large language models on our curated ESG dataset, using the methodology described in the previous section. Our evaluation includes models from different families and parameter sizes: glm-4.7-flash (30B MoE, q4\_K\_M), nemotron-3-nano:4b (4B), qwen3:4b-instruct (4B), gemma3:4b (4B), gemma4:e4b (8B, 4.5B effective), gemma4:e2b (2B, 2.3B effective), and ministral-3:8b (8B). This diverse model selection enables systematic analysis of performance characteristics across different architectures and sizes in the open-source ecosystem. The evaluation encompasses both retrieval and generation performance across environmental, social, and governance domains. To ensure consistency across all experiments, we use qwen3-embedding:4b for both dataset generation and RAG evaluation processes.

\subsection{Overall Performance Comparison}

Table~\ref{tab:rag_performance} presents the performance evaluation of seven open-source LLMs across seven RAGAS metrics. The results reveal clear differences in model capabilities across different aspects of the RAG pipeline, with notable performance differences emerging between model families and sizes.

\begin{table*}[htbp]
\centering
\caption{RAG Performance Evaluation Results Across Seven Open-Source LLMs for ESG Domain. LLMCP(Ref) = LLM Context Precision With Reference, CP = Context Precision, CR = Context Recall, CtxRel = Context Relevance, Faith = Faithfulness, AR = Answer Relevancy, FC = Factual Correctness (F1). Best score per column is in bold; ties highlight all tied models.}
\label{tab:rag_performance}
\resizebox{\textwidth}{!}{%
\begin{tabular}{l|cccc|cc|c}
\toprule
\multirow{2}{*}{\textbf{Model}} & \multicolumn{4}{c|}{\textbf{Retriever Metrics}} & \multicolumn{2}{c|}{\textbf{Generator Metrics}} & \textbf{Answer Quality} \\
\cmidrule{2-8}
 & \textbf{LLMCP(Ref)} & \textbf{CP} & \textbf{CR} & \textbf{CtxRel} & \textbf{Faith} & \textbf{AR} & \textbf{FC (F1)} \\
\midrule
glm-4.7-flash    & \textbf{0.878} & 0.803          & 0.600          & 0.985 & \textbf{0.822} & 0.818          & 0.387          \\
nemotron-3-nano  & \textbf{0.878} & 0.792          & \textbf{0.605} & 0.985 & 0.625          & 0.851          & 0.402          \\
qwen3            & 0.860          & \textbf{0.805} & \textbf{0.605} & 0.985 & 0.719          & 0.863          & \textbf{0.449} \\
gemma3           & 0.875          & \textbf{0.805} & \textbf{0.605} & 0.985 & 0.776          & 0.877          & 0.427          \\
gemma4:e4b       & 0.830          & 0.783          & 0.578          & 0.965 & 0.796          & 0.762          & 0.405          \\
gemma4:e2b       & 0.830          & 0.783          & 0.578          & 0.965 & 0.784          & 0.760          & 0.423          \\
ministral-3      & 0.875          & \textbf{0.805} & 0.600          & 0.985 & 0.607          & \textbf{0.881} & 0.405          \\
\bottomrule
\end{tabular}%
}
\end{table*}

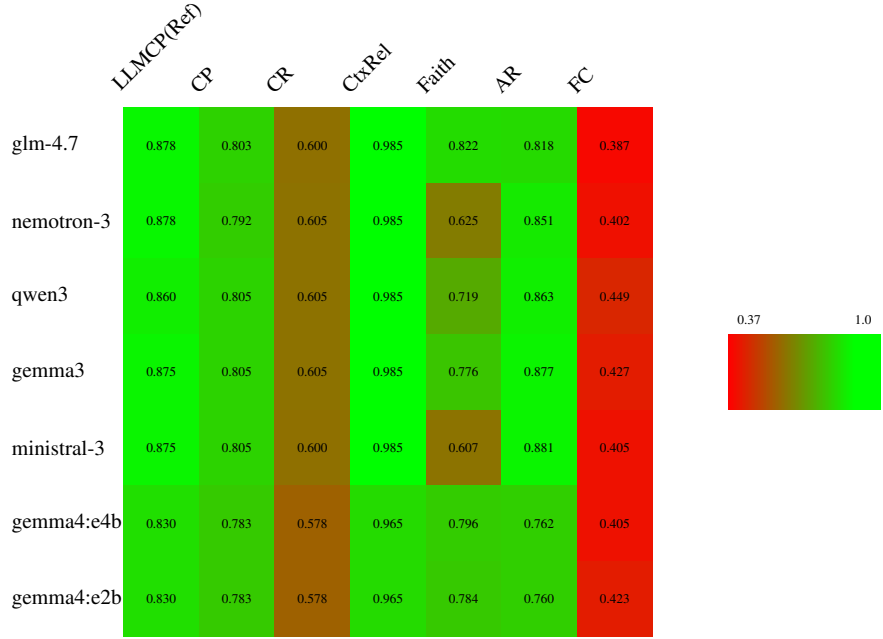
\begin{figure*}[htbp]
\centering
\caption{RAG performance heatmap across seven models and metrics. Darker green indicates higher scores; lighter/redder indicates lower scores. Retrieval metrics (LLMCP(Ref), CP, CR, CtxRel) cluster tightly, while generation metrics (Faith, AR, FC) show wider variance.}
\label{fig:rag_heatmap}
\begin{tikzpicture}

\pgfmathsetmacro{\r}{0.032}\pgfmathsetmacro{\g}{0.968} \definecolor{c0}{rgb}{\r,\g,0} \fill[fill=c0] (0,0) rectangle (1,-1); \node[font=\tiny] at (0.5,-0.5) {0.878};
\pgfmathsetmacro{\r}{0.175}\pgfmathsetmacro{\g}{0.825} \definecolor{c1}{rgb}{\r,\g,0} \fill[fill=c1] (1,0) rectangle (2,-1); \node[font=\tiny] at (1.5,-0.5) {0.803};
\pgfmathsetmacro{\r}{0.562}\pgfmathsetmacro{\g}{0.438} \definecolor{c2}{rgb}{\r,\g,0} \fill[fill=c2] (2,0) rectangle (3,-1); \node[font=\tiny] at (2.5,-0.5) {0.600};
\pgfmathsetmacro{\r}{0.000}\pgfmathsetmacro{\g}{1.000} \definecolor{c3}{rgb}{\r,\g,0} \fill[fill=c3] (3,0) rectangle (4,-1); \node[font=\tiny] at (3.5,-0.5) {0.985};
\pgfmathsetmacro{\r}{0.139}\pgfmathsetmacro{\g}{0.861} \definecolor{c4}{rgb}{\r,\g,0} \fill[fill=c4] (4,0) rectangle (5,-1); \node[font=\tiny] at (4.5,-0.5) {0.822};
\pgfmathsetmacro{\r}{0.147}\pgfmathsetmacro{\g}{0.853} \definecolor{c5}{rgb}{\r,\g,0} \fill[fill=c5] (5,0) rectangle (6,-1); \node[font=\tiny] at (5.5,-0.5) {0.818};
\pgfmathsetmacro{\r}{0.968}\pgfmathsetmacro{\g}{0.032} \definecolor{c6}{rgb}{\r,\g,0} \fill[fill=c6] (6,0) rectangle (7,-1); \node[font=\tiny] at (6.5,-0.5) {0.387};
\pgfmathsetmacro{\r}{0.032}\pgfmathsetmacro{\g}{0.968} \definecolor{c7}{rgb}{\r,\g,0} \fill[fill=c7] (0,-1) rectangle (1,-2); \node[font=\tiny] at (0.5,-1.5) {0.878};
\pgfmathsetmacro{\r}{0.196}\pgfmathsetmacro{\g}{0.804} \definecolor{c8}{rgb}{\r,\g,0} \fill[fill=c8] (1,-1) rectangle (2,-2); \node[font=\tiny] at (1.5,-1.5) {0.792};
\pgfmathsetmacro{\r}{0.552}\pgfmathsetmacro{\g}{0.448} \definecolor{c9}{rgb}{\r,\g,0} \fill[fill=c9] (2,-1) rectangle (3,-2); \node[font=\tiny] at (2.5,-1.5) {0.605};
\pgfmathsetmacro{\r}{0.000}\pgfmathsetmacro{\g}{1.000} \definecolor{c10}{rgb}{\r,\g,0} \fill[fill=c10] (3,-1) rectangle (4,-2); \node[font=\tiny] at (3.5,-1.5) {0.985};
\pgfmathsetmacro{\r}{0.514}\pgfmathsetmacro{\g}{0.486} \definecolor{c11}{rgb}{\r,\g,0} \fill[fill=c11] (4,-1) rectangle (5,-2); \node[font=\tiny] at (4.5,-1.5) {0.625};
\pgfmathsetmacro{\r}{0.084}\pgfmathsetmacro{\g}{0.916} \definecolor{c12}{rgb}{\r,\g,0} \fill[fill=c12] (5,-1) rectangle (6,-2); \node[font=\tiny] at (5.5,-1.5) {0.851};
\pgfmathsetmacro{\r}{0.939}\pgfmathsetmacro{\g}{0.061} \definecolor{c13}{rgb}{\r,\g,0} \fill[fill=c13] (6,-1) rectangle (7,-2); \node[font=\tiny] at (6.5,-1.5) {0.402};
\pgfmathsetmacro{\r}{0.066}\pgfmathsetmacro{\g}{0.934} \definecolor{c14}{rgb}{\r,\g,0} \fill[fill=c14] (0,-2) rectangle (1,-3); \node[font=\tiny] at (0.5,-2.5) {0.860};
\pgfmathsetmacro{\r}{0.171}\pgfmathsetmacro{\g}{0.829} \definecolor{c15}{rgb}{\r,\g,0} \fill[fill=c15] (1,-2) rectangle (2,-3); \node[font=\tiny] at (1.5,-2.5) {0.805};
\pgfmathsetmacro{\r}{0.552}\pgfmathsetmacro{\g}{0.448} \definecolor{c16}{rgb}{\r,\g,0} \fill[fill=c16] (2,-2) rectangle (3,-3); \node[font=\tiny] at (2.5,-2.5) {0.605};
\pgfmathsetmacro{\r}{0.000}\pgfmathsetmacro{\g}{1.000} \definecolor{c17}{rgb}{\r,\g,0} \fill[fill=c17] (3,-2) rectangle (4,-3); \node[font=\tiny] at (3.5,-2.5) {0.985};
\pgfmathsetmacro{\r}{0.335}\pgfmathsetmacro{\g}{0.665} \definecolor{c18}{rgb}{\r,\g,0} \fill[fill=c18] (4,-2) rectangle (5,-3); \node[font=\tiny] at (4.5,-2.5) {0.719};
\pgfmathsetmacro{\r}{0.061}\pgfmathsetmacro{\g}{0.939} \definecolor{c19}{rgb}{\r,\g,0} \fill[fill=c19] (5,-2) rectangle (6,-3); \node[font=\tiny] at (5.5,-2.5) {0.863};
\pgfmathsetmacro{\r}{0.850}\pgfmathsetmacro{\g}{0.150} \definecolor{c20}{rgb}{\r,\g,0} \fill[fill=c20] (6,-2) rectangle (7,-3); \node[font=\tiny] at (6.5,-2.5) {0.449};
\pgfmathsetmacro{\r}{0.040}\pgfmathsetmacro{\g}{0.960} \definecolor{c21}{rgb}{\r,\g,0} \fill[fill=c21] (0,-3) rectangle (1,-4); \node[font=\tiny] at (0.5,-3.5) {0.875};
\pgfmathsetmacro{\r}{0.171}\pgfmathsetmacro{\g}{0.829} \definecolor{c22}{rgb}{\r,\g,0} \fill[fill=c22] (1,-3) rectangle (2,-4); \node[font=\tiny] at (1.5,-3.5) {0.805};
\pgfmathsetmacro{\r}{0.552}\pgfmathsetmacro{\g}{0.448} \definecolor{c23}{rgb}{\r,\g,0} \fill[fill=c23] (2,-3) rectangle (3,-4); \node[font=\tiny] at (2.5,-3.5) {0.605};
\pgfmathsetmacro{\r}{0.000}\pgfmathsetmacro{\g}{1.000} \definecolor{c24}{rgb}{\r,\g,0} \fill[fill=c24] (3,-3) rectangle (4,-4); \node[font=\tiny] at (3.5,-3.5) {0.985};
\pgfmathsetmacro{\r}{0.226}\pgfmathsetmacro{\g}{0.774} \definecolor{c25}{rgb}{\r,\g,0} \fill[fill=c25] (4,-3) rectangle (5,-4); \node[font=\tiny] at (4.5,-3.5) {0.776};
\pgfmathsetmacro{\r}{0.043}\pgfmathsetmacro{\g}{0.957} \definecolor{c26}{rgb}{\r,\g,0} \fill[fill=c26] (5,-3) rectangle (6,-4); \node[font=\tiny] at (5.5,-3.5) {0.877};
\pgfmathsetmacro{\r}{0.892}\pgfmathsetmacro{\g}{0.108} \definecolor{c27}{rgb}{\r,\g,0} \fill[fill=c27] (6,-3) rectangle (7,-4); \node[font=\tiny] at (6.5,-3.5) {0.427};
\pgfmathsetmacro{\r}{0.040}\pgfmathsetmacro{\g}{0.960} \definecolor{c28}{rgb}{\r,\g,0} \fill[fill=c28] (0,-4) rectangle (1,-5); \node[font=\tiny] at (0.5,-4.5) {0.875};
\pgfmathsetmacro{\r}{0.171}\pgfmathsetmacro{\g}{0.829} \definecolor{c29}{rgb}{\r,\g,0} \fill[fill=c29] (1,-4) rectangle (2,-5); \node[font=\tiny] at (1.5,-4.5) {0.805};
\pgfmathsetmacro{\r}{0.562}\pgfmathsetmacro{\g}{0.438} \definecolor{c30}{rgb}{\r,\g,0} \fill[fill=c30] (2,-4) rectangle (3,-5); \node[font=\tiny] at (2.5,-4.5) {0.600};
\pgfmathsetmacro{\r}{0.000}\pgfmathsetmacro{\g}{1.000} \definecolor{c31}{rgb}{\r,\g,0} \fill[fill=c31] (3,-4) rectangle (4,-5); \node[font=\tiny] at (3.5,-4.5) {0.985};
\pgfmathsetmacro{\r}{0.549}\pgfmathsetmacro{\g}{0.451} \definecolor{c32}{rgb}{\r,\g,0} \fill[fill=c32] (4,-4) rectangle (5,-5); \node[font=\tiny] at (4.5,-4.5) {0.607};
\pgfmathsetmacro{\r}{0.040}\pgfmathsetmacro{\g}{0.960} \definecolor{c33}{rgb}{\r,\g,0} \fill[fill=c33] (5,-4) rectangle (6,-5); \node[font=\tiny] at (5.5,-4.5) {0.881};
\pgfmathsetmacro{\r}{0.933}\pgfmathsetmacro{\g}{0.067} \definecolor{c34}{rgb}{\r,\g,0} \fill[fill=c34] (6,-4) rectangle (7,-5); \node[font=\tiny] at (6.5,-4.5) {0.405};

\pgfmathsetmacro{\r}{0.125}\pgfmathsetmacro{\g}{0.875} \definecolor{c35}{rgb}{\r,\g,0} \fill[fill=c35] (0,-5) rectangle (1,-6); \node[font=\tiny] at (0.5,-5.5) {0.830};
\pgfmathsetmacro{\r}{0.207}\pgfmathsetmacro{\g}{0.793} \definecolor{c36}{rgb}{\r,\g,0} \fill[fill=c36] (1,-5) rectangle (2,-6); \node[font=\tiny] at (1.5,-5.5) {0.783};
\pgfmathsetmacro{\r}{0.619}\pgfmathsetmacro{\g}{0.381} \definecolor{c37}{rgb}{\r,\g,0} \fill[fill=c37] (2,-5) rectangle (3,-6); \node[font=\tiny] at (2.5,-5.5) {0.578};
\pgfmathsetmacro{\r}{0.143}\pgfmathsetmacro{\g}{0.857} \definecolor{c38}{rgb}{\r,\g,0} \fill[fill=c38] (3,-5) rectangle (4,-6); \node[font=\tiny] at (3.5,-5.5) {0.965};
\pgfmathsetmacro{\r}{0.200}\pgfmathsetmacro{\g}{0.800} \definecolor{c39}{rgb}{\r,\g,0} \fill[fill=c39] (4,-5) rectangle (5,-6); \node[font=\tiny] at (4.5,-5.5) {0.796};
\pgfmathsetmacro{\r}{0.187}\pgfmathsetmacro{\g}{0.813} \definecolor{c40}{rgb}{\r,\g,0} \fill[fill=c40] (5,-5) rectangle (6,-6); \node[font=\tiny] at (5.5,-5.5) {0.762};
\pgfmathsetmacro{\r}{0.933}\pgfmathsetmacro{\g}{0.067} \definecolor{c41}{rgb}{\r,\g,0} \fill[fill=c41] (6,-5) rectangle (7,-6); \node[font=\tiny] at (6.5,-5.5) {0.405};

\pgfmathsetmacro{\r}{0.125}\pgfmathsetmacro{\g}{0.875} \definecolor{c42}{rgb}{\r,\g,0} \fill[fill=c42] (0,-6) rectangle (1,-7); \node[font=\tiny] at (0.5,-6.5) {0.830};
\pgfmathsetmacro{\r}{0.207}\pgfmathsetmacro{\g}{0.793} \definecolor{c43}{rgb}{\r,\g,0} \fill[fill=c43] (1,-6) rectangle (2,-7); \node[font=\tiny] at (1.5,-6.5) {0.783};
\pgfmathsetmacro{\r}{0.619}\pgfmathsetmacro{\g}{0.381} \definecolor{c44}{rgb}{\r,\g,0} \fill[fill=c44] (2,-6) rectangle (3,-7); \node[font=\tiny] at (2.5,-6.5) {0.578};
\pgfmathsetmacro{\r}{0.143}\pgfmathsetmacro{\g}{0.857} \definecolor{c45}{rgb}{\r,\g,0} \fill[fill=c45] (3,-6) rectangle (4,-7); \node[font=\tiny] at (3.5,-6.5) {0.965};
\pgfmathsetmacro{\r}{0.210}\pgfmathsetmacro{\g}{0.790} \definecolor{c46}{rgb}{\r,\g,0} \fill[fill=c46] (4,-6) rectangle (5,-7); \node[font=\tiny] at (4.5,-6.5) {0.784};
\pgfmathsetmacro{\r}{0.190}\pgfmathsetmacro{\g}{0.810} \definecolor{c47}{rgb}{\r,\g,0} \fill[fill=c47] (5,-6) rectangle (6,-7); \node[font=\tiny] at (5.5,-6.5) {0.760};
\pgfmathsetmacro{\r}{0.892}\pgfmathsetmacro{\g}{0.108} \definecolor{c48}{rgb}{\r,\g,0} \fill[fill=c48] (6,-6) rectangle (7,-7); \node[font=\tiny] at (6.5,-6.5) {0.423};

\node[right, font=\footnotesize] at (-1.6, -0.5) {glm-4.7};
\node[right, font=\footnotesize] at (-1.6, -1.5) {nemotron-3};
\node[right, font=\footnotesize] at (-1.6, -2.5) {qwen3};
\node[right, font=\footnotesize] at (-1.6, -3.5) {gemma3};
\node[right, font=\footnotesize] at (-1.6, -4.5) {ministral-3};
\node[right, font=\footnotesize] at (-1.6, -5.5) {gemma4:e4b};
\node[right, font=\footnotesize] at (-1.6, -6.5) {gemma4:e2b};

\node[rotate=45, anchor=south west, font=\footnotesize] at (0.0, 0.0) {LLMCP(Ref)};
\node[rotate=45, anchor=south west, font=\footnotesize] at (1.0, 0.0) {CP};
\node[rotate=45, anchor=south west, font=\footnotesize] at (2.0, 0.0) {CR};
\node[rotate=45, anchor=south west, font=\footnotesize] at (3.0, 0.0) {CtxRel};
\node[rotate=45, anchor=south west, font=\footnotesize] at (4.0, 0.0) {Faith};
\node[rotate=45, anchor=south west, font=\footnotesize] at (5.0, 0.0) {AR};
\node[rotate=45, anchor=south west, font=\footnotesize] at (6.0, 0.0) {FC};

\foreach \x in {0,...,100} {
    \pgfmathsetmacro{\v}{0.37 + \x/100 * 0.63}
    \pgfmathsetmacro{\rr}{max(0, 1 - (\v-0.37)/(0.63)*1.2)}
    \pgfmathsetmacro{\gg}{max(0, (\v-0.37)/(0.63)*1.2)}
    \definecolor{barcolor}{rgb}{\rr,\gg,0}
    \fill[fill=barcolor] ({8 + \x*0.02}, {-4}) rectangle ({8 + (\x+1)*0.02}, {-3});
}
\node[anchor=west, font=\tiny] at (8, -2.8) {0.37};
\node[anchor=east, font=\tiny] at (10.02, -2.8) {1.0};

\end{tikzpicture}
\end{figure*}

\subsection{Retrieval Performance Analysis}
\label{sec:results_retrieval}

Retrieval performance is strong but no longer perfect once judged with the RAGAS LLM-based metrics. Context relevance reaches 0.965--0.985 across all models, indicating the top-1 retrieved chunk is almost always topically aligned with the query. LLM-judged context precision with reference is also high (0.830--0.878), confirming that the retrieved chunk usually contains the information needed to derive the reference answer. Context precision and context recall sit at 0.783--0.805 and 0.578--0.605 respectively.

The shared retriever (qwen3-embedding:4b, top-$k$=1) means CtxRel would be identical across models in a deterministic retrieval setup; the small differences in LLMCP(Ref), CP, CR, and CtxRel arise from the LLM judge's per-sample interpretation of whether the retrieved chunk fully supports the reference. The Gemma 4 models (e4b and e2b) show slightly lower retrieval-side scores compared to the other five models, which may reflect differences in how their responses interact with the LLM judge's scoring criteria. Recall below unity reflects a realistic ceiling for top-$k$=1 retrieval: when the reference answer spans information that overlaps multiple chunks, a single retrieved chunk cannot cover all of it. These scores are not driven by an embedding-model mismatch between retrieval and judging --- both stages use \texttt{qwen3-embedding:4b} --- but rather by genuine retrieval gaps that the LLM judge surfaces.

\subsection{Generation Performance Analysis}

Generator performance reveals substantial differences between models, particularly in faithfulness. Faithfulness shows the widest spread, with glm-4.7-flash:q4\_K\_M leading (0.822), followed by gemma4:e4b (0.796), gemma4:e2b (0.784), gemma3:4b (0.776), qwen3:4b-instruct (0.719), nemotron-3-nano:4b (0.625), and ministral-3:8b (0.607). The 0.215 spread between best and worst on this metric is the dominant generation-quality differentiator. Notably, both Gemma 4 models outperform gemma3:4b on faithfulness, with gemma4:e4b (0.796) ranking second overall, despite having lower retrieval-side scores. Since all seven models receive the same anchoring system prompt and the same retrieved context (Section~\ref{sec:prompt_format}), this spread is consistent with model-level differences in how strictly each generator follows the ``answer from documents only'' instruction, though some portion of the variation may also reflect unnormalized default decoding settings in Step 2.

Answer relevancy shows a wider spread when Gemma 4 models are included (0.760--0.881). The Gemma 4 models score notably lower on this metric (gemma4:e4b 0.762, gemma4:e2b 0.760), while the remaining five models cluster tightly (0.818--0.881) with ministral-3:8b first (0.881), followed by gemma3:4b (0.877), qwen3:4b-instruct (0.863), nemotron-3-nano:4b (0.851), and glm-4.7-flash:q4\_K\_M (0.818). The lower AR scores for Gemma 4 models suggest their responses may be less directly aligned with the user query, possibly due to differences in instruction-following behavior or decoding defaults.

\subsection{Factual Correctness and Domain-Specific Performance}

Factual correctness scores range from 0.387 to 0.449 across evaluated models. Qwen3:4b-instruct leads (0.449), followed by gemma3:4b (0.427), gemma4:e2b (0.423), ministral-3:8b (0.405), gemma4:e4b (0.405), nemotron-3-nano:4b (0.402), and glm-4.7-flash:q4\_K\_M (0.387). The 0.062 spread is much narrower than the faithfulness spread, suggesting all seven models face similar headwinds when extracting specific ESG facts from a single retrieved chunk. Even the top performer leaves substantial headroom for improvement.

\subsubsection{ESG Domain Complexity Challenges}

The factual correctness challenges in ESG applications stem from several domain-specific complexities that distinguish sustainability reporting from general knowledge domains \cite{berg2021esg}:

\begin{itemize}
    \item \textbf{Technical Terminology Evolution}: ESG reporting terminology evolves rapidly with new frameworks (GRI, SASB, TCFD, ESRS) and regulatory requirements, creating challenges for models trained on historical data \cite{eu2023csrd}.
    
    \item \textbf{Quantitative Data Precision}: ESG metrics require high numerical precision (e.g., carbon emissions in tons, percentage reductions, monetary values) where small errors can have significant regulatory implications \cite{carney2015transition}.
    
    \item \textbf{Contextual Dependencies}: ESG facts often depend on specific regulatory contexts, industry standards, and geographic variations that general-purpose models may not capture accurately \cite{berg2021esg,EU_CSRD_2022,ESRS2023}.
    
    \item \textbf{Forward-Looking Statements}: ESG reports contain numerous forward-looking commitments and targets that require careful distinction between historical facts and projected goals \cite{tcfd2017recommendations}.
\end{itemize}

\subsubsection{Model Family Performance Patterns}

The variation in factual correctness across model families may reflect differences in training emphasis and data composition, though the precise causes cannot be determined from evaluation scores alone:

\begin{itemize}
    \item \textbf{Qwen3:4b-instruct (0.449)}: Top factual correctness; given the self-enhancement caveat (Section~\ref{sec:limitations}), this leadership should be interpreted cautiously, as the dataset generator and RAGAS judge are also instances of qwen3:4b-instruct~\cite{qwen3tech}.

    \item \textbf{Gemma3:4b (0.427)}: Strong factual performance; the underlying reasons are unclear from this evaluation alone~\cite{team2024gemma}.

    \item \textbf{Gemma4:e2b (0.423)}: Strong factual correctness for a 2B model, nearly matching gemma3:4b despite having fewer parameters~\cite{team2024gemma}.

    \item \textbf{Gemma4:e4b (0.405)}: Mid-range factual correctness; performs similarly to ministral-3:8b despite being a comparable effective parameter count~\cite{team2024gemma}.

    \item \textbf{Ministral-3:8b (0.405)}: Mid-range factual correctness paired with the highest answer relevancy (0.881)~\cite{ministral3}.

    \item \textbf{Nemotron-3-nano:4b (0.402)}: Mid-range factual correctness; near-tied with ministral-3 and gemma4:e4b~\cite{nemotron3nano}.

    \item \textbf{GLM-4.7-flash:q4\_K\_M (0.387)}: Lowest factual correctness; the 4-bit quantization is a plausible contributor, though the cause cannot be isolated from this evaluation alone~\cite{zhipu2024glm,frantar2023qlora}.
\end{itemize}

\subsubsection{Practical Implications and Improvement Strategies}

The factual correctness results have important implications for ESG RAG deployment:

\begin{itemize}
    \item \textbf{Human-in-the-Loop Validation}: Current performance levels suggest that critical ESG applications should incorporate human validation mechanisms for high-stakes decisions \cite{amodei2016concrete}.
    
    \item \textbf{Domain-Specific Fine-Tuning}: The performance variation across models indicates significant potential for improvement through ESG-specific fine-tuning using curated sustainability datasets \cite{susgen2025,ESGCID25}.
    
    \item \textbf{Hybrid Retrieval Architectures}: Larger top-$k$, rerankers, and modular retrieval pipelines could leverage complementary retrieval signals to improve overall factual accuracy \cite{gao2023rag}.
    
    \item \textbf{Retrieval Enhancement}: Improved retrieval systems with better ESG-specific indexing could provide more accurate context, reducing factual errors \cite{birti2025esg}.
\end{itemize}

The narrow 0.387--0.449 band across all seven models, all sitting well below 0.5, indicates that factual correctness remains the weakest link of the pipeline in this evaluation. This pattern persists under a fixed retrieval configuration with relatively strong retrieval-side scores (LLMCP(Ref) up to 0.878) and high answer relevancy ($\geq$0.760 for every model), indicating that factual correctness remains a major bottleneck in the current pipeline configuration. Because retrieval was fixed at top-$k$=1 and no retrieval ablation was performed, these results should not be interpreted as ruling out retrieval improvements as a path to better factual accuracy. ESG applications that demand high precision for regulatory compliance and stakeholder trust would therefore benefit substantially from domain-specific fine-tuning or human-in-the-loop validation, regardless of which generator is chosen from this set.

\subsection{Model-Specific Insights}

\textbf{glm-4.7-flash:q4\_K\_M} leads faithfulness (0.822) and ties for the highest LLM-judged context precision with reference (0.878). Its trade-off is the lowest answer relevancy (0.818) and lowest factual correctness (0.387) of the seven, plausibly influenced by 4-bit quantization. Within this evaluation setting, it appears best suited to faithfulness-critical applications where staying anchored to retrieved evidence outweighs phrasing fluency.

\textbf{nemotron-3-nano:4b} ties for the highest LLMCP(Ref) (0.878) and ranks mid-pack on AR (0.851) and FC (0.402), but posts the lowest faithfulness (0.625) of the set. Within this evaluation setting, it appears to be a solid generalist for retrieval-side use cases, but weaker when strict context grounding is required.

\textbf{qwen3:4b-instruct} leads factual correctness (0.449) and posts competitive faithfulness (0.719) and answer relevancy (0.863). Because the dataset generator and RAGAS judge are also qwen3 family, this leadership is qualified by the self-enhancement caveat (Section~\ref{sec:limitations}).

\textbf{gemma3:4b} is the most balanced model: second on faithfulness (0.776), second on answer relevancy (0.877), and second on factual correctness (0.427) --- no clear weakness on any generation metric. Within this evaluation setting, it is a reasonable default when the deployment requirements are not yet known.

\textbf{gemma4:e4b} (8B, 4.5B effective) ranks second on faithfulness (0.796) but posts lower answer relevancy (0.762) than the other five models, with mid-range factual correctness (0.405). Its strong faithfulness suggests effective instruction-following for context grounding, but the lower AR indicates room for improvement in query alignment. Within this evaluation setting, it is a strong candidate for faithfulness-prioritizing applications.

\textbf{gemma4:e2b} (2B, 2.3B effective) is noteworthy for its efficiency: despite being the smallest model in the set, it achieves competitive faithfulness (0.784) and factual correctness (0.423). It trails only on answer relevancy (0.760), where it is the lowest of the set. Within this evaluation setting, it is a compelling choice for resource-constrained deployments that still require strong context grounding.

\textbf{ministral-3:8b} leads answer relevancy (0.881) but posts the lowest faithfulness (0.607) of the set, with mid-range factual correctness (0.405). Within this evaluation setting, it appears best suited to user-facing applications where on-topic phrasing matters most and the consequence of mild context drift is acceptable.

Overall, no single model dominates every metric. Model selection should be driven by which generation property the application weighs most heavily, with the self-enhancement caveat in mind when comparing qwen3 against the others.

\subsection{Implications for ESG RAG Applications}

Within the reported ESG RAG evaluation setting, glm-4.7-flash:q4\_K\_M is the strongest choice for faithfulness-critical tasks (0.822), where the system must stay anchored to the retrieved evidence rather than draw on parametric knowledge. For applications prioritizing factual accuracy, qwen3:4b-instruct offers the highest factual correctness (0.449) and competitive answer relevancy (0.863), making it well suited for compliance-focused ESG systems --- subject to the self-enhancement caveat in Section~\ref{sec:limitations}. For applications prioritizing query-response alignment, ministral-3:8b delivers the highest answer relevancy (0.881), making it suitable for user-facing ESG information systems. Gemma3:4b offers the most balanced profile across the three generation metrics in this evaluation and is a reasonable default when deployment requirements are not yet known. Among the Gemma 4 family, gemma4:e4b (8B) is a strong faithfulness-oriented alternative (0.796) with the edge-deployable gemma4:e2b (2B) achieving surprisingly competitive factual correctness (0.423) at minimal computational cost, making it suited for resource-constrained settings.

\section{Limitations}
\label{sec:limitations}

Several limitations should be considered when interpreting the findings of this study.

\textbf{Corpus scope}: The evaluation corpus comprises 498 ESG reports from EU-listed companies, collected from publicly available corporate websites. Findings may not generalize to ESG reporting in other geographic regions, languages, or regulatory frameworks.

\textbf{Synthetic evaluation set}: The 100 QA pairs used for evaluation were generated synthetically by the same LLM family used for dataset construction (qwen3:4b-instruct). Real user queries may differ substantially in phrasing, complexity, and information needs, which could affect the observed performance rankings.

\textbf{Subset selection}: The reported results are based on a deterministic prefix of the 284-pair synthetic testset rather than a randomized or stratified evaluation subset. The observed rankings may therefore be sensitive to subset selection.

\textbf{Top-$k$=1 retrieval ceiling}: All retrieval results are bounded by single-chunk top-1 retrieval. Context recall around 0.60 reflects cases where the reference answer's supporting evidence spans information beyond a single 1000-character chunk. Retrieval performance under higher top-$k$, larger chunk sizes, or independent (non-synthetic) query sets is unknown.

\textbf{No non-RAG baseline}: This study evaluates RAG pipelines only; direct LLM inference without retrieval was not included. The absolute benefit of the retrieval component therefore cannot be quantified from these results alone.

\textbf{No hyperparameter ablation}: Chunk size (1000 characters), overlap (200 characters), and top-k (1) were fixed throughout. The sensitivity of results to these choices was not studied.

\textbf{LLM-as-judge variability}: RAGAS metrics rely on LLM-based scoring. In our setup, scoring is mediated by the RAGAS LLM wrapper rather than a raw direct call to the judge model, which applies near-deterministic single-completion behavior. Even so, judge-model choice and LLM-as-judge effects may still introduce uncertainty not captured by point estimates alone.

\textbf{Generator decoding control}: Step 2 generation used direct \texttt{ChatOllama(model=...)} calls without explicitly normalizing temperature or related sampling parameters across models. As a result, some observed differences may reflect both checkpoint differences and model-specific default decoding behavior.

\textbf{Sampling fraction}: Only 300 of 258{,}590 corpus chunks (0.12\%) were drawn for QA generation, covering 181 of the 498 source documents. Results reflect performance on this sample rather than full-corpus coverage.

\textbf{Self-enhancement risk}: The dataset generator, one of the evaluated models, and the RAGAS judge are all instances of \texttt{qwen3:4b-instruct}. Although the generator/judge call paths are separate from inference, qwen3 scores on faithfulness, answer relevancy, and factual correctness may be inflated relative to other models due to stylistic and lexical alignment between generation, response, and scoring~\cite{bedrock_synth_rag2024}. Cross-validation with an independent judge model is left to future work.

\section{Conclusion and Future Work}

This study presents an empirical evaluation framework for open-source Large Language Models (LLMs) in ESG (Environmental, Social, and Governance) KPI extraction using Retrieval-Augmented Generation (RAG). Our analysis evaluates seven open-source models ranging from 2B to 30B parameters on a dataset of 498 ESG documents, revealing marked differences in performance characteristics across model architectures and sizes.

The performance evaluation reveals genuine variation across both retrieval and generation. Retrieval is strong but not perfect (context recall $\approx$0.58--0.61, context precision $\approx$0.78--0.81, context relevance 0.965--0.985, LLM-judged context precision with reference $\approx$0.83--0.88), bounded by the top-$k$=1 design. Generation diverges most on faithfulness (0.607--0.822) and least on answer relevancy (0.760--0.881). Different models lead different metrics: glm-4.7-flash:q4\_K\_M leads faithfulness (0.822), qwen3:4b-instruct leads factual correctness (0.449), and ministral-3:8b leads answer relevancy (0.881). Factual correctness across all models (0.387--0.449) indicates ongoing opportunities for domain-specific fine-tuning and further enhancement.

Strategically, this evaluation framework offers ESG practitioners directional guidance for open-source model selection based on their specific requirements. Under the reported configuration, organizations can use these results as directional guidance: glm-4.7-flash:q4\_K\_M for faithfulness-critical applications, qwen3:4b-instruct for factual-accuracy-critical tasks (subject to the self-enhancement caveat), ministral-3:8b for query-focused applications, gemma4:e4b as a faithfulness-oriented alternative, gemma4:e2b for resource-constrained deployments, and gemma3:4b as a balanced default. The narrow factual-correctness band across all seven models indicates that domain-specific fine-tuning and human-in-the-loop validation remain valuable directions regardless of generator choice.

The study establishes a transparent and reproducible methodology for evaluating open-source LLMs in sustainability reporting, providing essential guidance for organizations seeking open-source AI solutions for ESG analytics.

Beyond the model comparison itself, the study contributes an ESG-RAG evaluation resource that combines a processed report corpus, grounded synthetic QA pairs, and reproducible evaluation artifacts for future benchmarking and ablation studies.

\subsection{Future Work}

Several avenues for future research emerge from this study:

\textbf{Extended Open-Source Model Benchmarking}: We plan to expand evaluation to include additional open-source models such as Llama 3, Mixtral, and emerging models from different parameter ranges (1B-70B). This broader comparison will provide deeper insights into the evolving landscape of open-source LLM capabilities for ESG applications.

\textbf{Domain-Specific Fine-Tuning}: Given the improved but still moderate factual correctness scores across all models, future work will investigate domain-specific fine-tuning approaches for ESG applications. This includes training models on specialized ESG datasets and developing adaptation techniques to further enhance factual accuracy in sustainability reporting contexts, building on the promising progress demonstrated in this evaluation.

\textbf{Broader Corpus Validation}: While our 498-document dataset provides coverage of EU companies' ESG reporting, validating our findings across diverse geographies, industries, and reporting frameworks will enhance the generalizability of the evaluation framework.

\textbf{Cost-Performance Optimization}: Future research will explore optimization techniques for open-source model deployment, including quantization strategies, model compression, and efficient serving architectures to maximize performance efficiency for ESG applications.

\textbf{Ablation Studies and Baselines}: This study fixed chunk size (1000 characters), overlap (200 characters), and top-k (1) without ablating their individual effects. Future work should vary these hyperparameters systematically and include a non-RAG baseline (direct LLM inference without retrieval) to quantify the retrieval component's contribution to overall performance.

\textbf{Hybrid Architecture Evaluation}: Future work will investigate hybrid approaches that combine different open-source models for different pipeline components, potentially optimizing both accuracy and computational efficiency for ESG RAG systems.

\section*{Acknowledgements}
\label{acknowledgements}
This research was partially supported by the Italian Research Center on High Performance Computing, Big Data and Quantum Computing (ICSC), funded by the European Union through NextGenerationEU (PNRR-HPC, CUP: C83C22000560007).

\printbibliography

@article{lewis2020retrieval,
  title={Retrieval-augmented generation for knowledge-intensive {NLP} tasks},
  author={Lewis, Patrick and Perez, Ethan and Piktus, Aleksandra and Petroni, Fabio and Karpukhin, Vladimir and Goyal, Naman and K{\"u}ttler, Heinrich and Lewis, Mike and Yih, Wen-tau and Rockt{\"a}schel, Tim and others},
  journal={Advances in Neural Information Processing Systems},
  volume={33},
  pages={9459--9474},
  year={2020}
}

@article{brown2020language,
  title={Language models are few-shot learners},
  author={Brown, Tom and Mann, Benjamin and Ryder, Nick and Subbiah, Melanie and Kaplan, Jared D and Dhariwal, Prafulla and Neelakantan, Arvind and Shyam, Pranav and Sastry, Girish and Askell, Amanda and others},
  journal={Advances in Neural Information Processing Systems},
  volume={33},
  pages={1877--1901},
  year={2020}
}

@article{meta2023llama,
  title={{LLaMA}: Open and efficient foundation language models},
  author={Touvron, Hugo and Lavril, Thibaut and Izacard, Gautier and Martinet, Xavier and Lachaux, Marie-Anne and Lacroix, Timoth{\'e}e and Rozi{\`e}re, Baptiste and Goyal, Naman and Hambro, Eric and Azhar, Faisal and others},
  journal={arXiv preprint arXiv:2302.13971},
  year={2023}
}

@article{ministral3,
  title={Ministral 3},
  author={Liu, Alexander H and others},
  journal={arXiv preprint arXiv:2601.08584},
  year={2026}
}

@misc{team2024gemma,
  title={Gemma: Open models based on {Gemini} research and technology},
  author={{Gemma Team, Google DeepMind}},
  howpublished={arXiv preprint arXiv:2403.08295},
  year={2024}
}

@misc{qwen3tech,
  title={{Qwen3} Technical Report},
  author={{Qwen Team}},
  howpublished={arXiv preprint arXiv:2505.09388},
  year={2025}
}

@misc{nemotron3nano,
  title={{Nemotron 3 Nano}: Open, Efficient Mixture-of-Experts Hybrid {Mamba}-{Transformer} Model for {Agentic} Reasoning},
  author={{NVIDIA}},
  howpublished={arXiv preprint arXiv:2512.20848},
  year={2025}
}

@misc{zhipu2024glm,
  title={{ChatGLM}: A family of large language models from {GLM}-130{B} to {GLM}-4 All Tools},
  author={{Team GLM} and Zeng, Aohan and Xu, Bin and Wang, Bowen and Zhang, Chenhui and Yin, Da and others},
  howpublished={arXiv preprint arXiv:2406.12793},
  year={2024}
}

@article{frantar2023qlora,
  title={{GPTQ}: Accurate post-training quantization for generative pre-trained transformers},
  author={Frantar, Elias and Ashkboos, Saleh and Hoefler, Torsten and Alistarh, Dan},
  journal={arXiv preprint arXiv:2210.17323},
  year={2022}
}

@misc{openllmleaderboard,
  title={Open {LLM} Leaderboard},
  author={Beeching, Edward and Fourrier, Cl{\'e}mentine and Habib, Nathan and Han, Sheon and Lambert, Nathan and Rajani, Nazneen and Sanseviero, Omar and Tunstall, Lewis and Wolf, Thomas},
  howpublished={\url{https://huggingface.co/spaces/HuggingFaceH4/open_llm_leaderboard}},
  year={2023},
  note={HuggingFace}
}

@article{amodei2016concrete,
  title={Concrete problems in {AI} safety},
  author={Amodei, Dario and Olah, Chris and Steinhardt, Jacob and Christiano, Paul and Schulman, John and Man{\'e}, Dan},
  journal={arXiv preprint arXiv:1606.06565},
  year={2016}
}

@article{berg2021esg,
  title={Aggregate confusion: The divergence of {ESG} ratings},
  author={Berg, Florian and K{\"o}lbel, Julian F and Rigobon, Roberto},
  journal={Review of Finance},
  volume={26},
  number={6},
  pages={1315--1344},
  year={2022}
}

@misc{carney2015transition,
  title={Breaking the tragedy of the horizon: Climate change and financial stability},
  author={Carney, Mark},
  howpublished={Speech at Lloyd's of London, Bank of England},
  year={2015}
}

@misc{tcfd2017recommendations,
  title={Recommendations of the {Task Force on Climate-related Financial Disclosures}},
  author={{Task Force on Climate-related Financial Disclosures}},
  howpublished={\url{https://www.fsb-tcfd.org/recommendations/}},
  year={2017}
}

@misc{eu2023csrd,
  title={{Corporate Sustainability Reporting Directive (CSRD)}},
  author={{European Commission}},
  howpublished={\url{https://finance.ec.europa.eu/capital-markets-union-and-financial-markets/company-reporting-and-auditing/company-reporting/corporate-sustainability-reporting_en}},
  year={2023}
}

@misc{bedrock_synth_rag2024,
  author       = {{Amazon Web Services}},
  title        = {Generate Synthetic Data for Evaluating {RAG} Systems Using {Amazon Bedrock}},
  howpublished = {\url{https://aws.amazon.com/blogs/machine-learning/generate-synthetic-data-for-evaluating-rag-systems-using-amazon-bedrock/}},
  year         = {2024},
  note         = {AWS Machine Learning Blog},
}

@article{rusu2024sustainability,
  title={Sustainability Performance Reporting.},
  author={Rusu, Teodora Maria and Odagiu, Antonia and Pop, Horia and Paulette, Laura},
  journal={Sustainability (2071-1050)},
  volume={16},
  number={19},
  year={2024}
}

@article{orsolin2024esg,
  title={Analysis of Indicators for the Integration, Implementation, and Development of the Environmental, Social, and Governance – ESG Report in ISE/B3 Companies},
  author={Orsolin, Augusto Londero and Ávila, Lucas Veiga and Trevisan, Marcelo and Dal Moro, Leila and Cavalcante, Diego Marques},
  journal={Revista Catarinense da Ciência Contábil},
  year={2024},
  doi={10.16930/2237-7662202435152}
}

@misc{cdp2025,
  title        = {{CDP: Turning Transparency to Action}},
  author       = {{CDP (Carbon Disclosure Project)}},
  howpublished = {\url{https://www.cdp.net/}},
  year         = {2025},
  note         = {Accessed: 2025-07-08},
}

@misc{GRI2021,
  title        = {{GRI Standards: Universal, Topic and Sector Standards}},
  author       = {{Global Reporting Initiative (GRI)}},
  howpublished = {\url{https://www.globalreporting.org/}},
  note         = {Framework first published October 2021, effective for reports in January 2023},
  year         = {2021},
  urldate      = {2025-07-08},
}

@misc{EU_CSRD_2022,
  title        = {{Directive (EU) 2022/2464 – Corporate Sustainability Reporting Directive (CSRD)}},
  author       = {{European Union, European Parliament \& Council}},
  howpublished = {\url{https://eur-lex.europa.eu/eli/dir/2022/2464/oj}},
  year         = {2022},
  note         = {Adopted 16 December 2022; in force 5 January 2023; phased reporting from January 2024 onwards},
  urldate      = {2025-07-08},
}

@misc{SASB2017,
  title        = {{Sustainability Accounting Standards Board (SASB) Conceptual Framework}},
  author       = {{Sustainability Accounting Standards Board (SASB)}},
  howpublished = {\url{https://www.sasb.org/}},
  year         = {2017},
  note         = {Standards launched for 77–79 industries; stewardship transferred to IFRS Foundation’s ISSB in August 2022},
  urldate      = {2025-07-08},
}

@misc{TCFD2017,
  title        = {{Recommendations of the Task Force on Climate‑related Financial Disclosures}},
  author       = {{Task Force on Climate‑related Financial Disclosures, Financial Stability Board}},
  howpublished = {\url{https://www.fsb-tcfd.org/}},
  year         = {2017},
  note         = {Final recommendations released June 2017; framework active until October 12, 2023; site maintained as resource; monitoring of progress handed to IFRS Foundation as of November 2023},
  urldate      = {2025-07-08},
}

@misc{ragas2024,
  author       = {ExplodingGradients},
  title        = {Ragas: Supercharge Your LLM Application Evaluations},
  year         = {2024},
  howpublished = {\url{https://github.com/explodinggradients/ragas}},
}

@INPROCEEDINGS{RAGTR,
  author={Ardic, Ozgur and Ozturk, Mahiye Uluyagmur and Demirtas, Irem and Arslan, Secil},
  booktitle={2024 32nd Signal Processing and Communications Applications Conference (SIU)}, 
  title={Information Extraction from Sustainability Reports in Turkish through RAG Approach}, 
  year={2024},
  volume={},
  number={},
  pages={1-4},
  doi={10.1109/SIU61531.2024.10600994}}

@article{ESGenius25,
  title={ESGenius: Benchmarking LLMs on Environmental, Social, and Governance (ESG) and Sustainability Knowledge},
  author={He, Chaoyue and Zhou, Xin and Wu, Yi and Yu, Xinjia and Zhang, Yan and Zhang, Lei and Wang, Di and Lyu, Shengfei and Xu, Hong and Wang, Xiaoqiao and others},
  journal={arXiv preprint arXiv:2506.01646},
  year={2025}
}

@article{MMESGBench25,
  title={Benchmarking Multimodal Understanding and Complex Reasoning for ESG Tasks},
  author={Zhang, Lei and Zhou, Xin and He, Chaoyue and Wang, Di and Wu, Yi and Xu, Hong and Liu, Wei and Miao, Chunyan},
  journal={arXiv preprint arXiv:2507.18932},
  year={2025}
}

@InProceedings{ESGConsultant25,
author="Ontiveros, Angel
and Nikishina, Irina
and Gomm, Moritz
and Schmitt, Christopher
and Biemann, Chris",
editor="Ichise, Ryutaro",
title="ESG-Consultant: Developing of an ESG Compliance Consulting Tool for Companies Using RAG",
booktitle="Natural Language Processing and Information Systems",
year="2026",
publisher="Springer Nature Switzerland",
address="Cham",
pages="223--229",
isbn="978-3-031-97144-0"
}

@article{ESGCID25,
  title={Enhancing Retrieval for ESGLLM via ESG-CID--A Disclosure Content Index Finetuning Dataset for Mapping GRI and ESRS},
  author={Ahmed, Shafiuddin Rehan and Shah, Ankit Parag and Tran, Quan Hung and Khetan, Vivek and Kang, Sukryool and Mehta, Ankit and Bao, Yujia and Wei, Wei},
  journal={arXiv preprint arXiv:2503.10674},
  year={2025}
}

@article{gao2023rag,
  author={Yunfan Gao and Yun Xiong and Xinyu Gao and Kangxiang Jia and Jinliu Pan and Yuxi Bi and Yi Dai and Jiawei Sun and Meng Wang and Haofen Wang},
  title={Retrieval-Augmented Generation for Large Language Models: A Survey},
  journal={arXiv preprint arXiv:2312.10997},
  year={2023}
}

@article{esgreveal2025,
  author={Yi Zou and Mengying Shi and Zhongjie Chen and Ye Liu and Jiayuan Chen and Siqiao Xue and Philip S. Yu and Fan Zhang},
  title={{ESGReveal}: An {LLM}-based Approach for Extracting Structured Data from {ESG} Reports},
  journal={Journal of Cleaner Production},
  volume={489},
  year={2025},
  doi={10.1016/j.jclepro.2024.144572}
}

@inproceedings{susgen2025,
  author={Qilong Wu and Xiaoneng Xiang and Hejia Huang and Xuan Wang and Ranjan Satapathy and Ricardo Shirota Filho and Bharadwaj Veeravalli},
  title={{SusGen-GPT}: A Data-Centric {LLM} for Financial {NLP} and Sustainability Report Generation},
  booktitle={Findings of the Association for Computational Linguistics: NAACL 2025},
  pages={1184--1203},
  year={2025},
  doi={10.18653/v1/2025.findings-naacl.66}
}

@inproceedings{birti2025esg,
  author={Mattia Birti and Francesco Osborne and Andrea Maurino},
  title={Optimizing Large Language Models for {ESG} Activity Detection in Financial Texts},
  booktitle={Proceedings of the 6th ACM International Conference on AI in Finance},
  year={2025},
  doi={10.1145/3768292.3770371}
}

@article{climatefinbench2025,
  author={Rafik Mankour and Yassine Chafai and Hamada Saleh and Ghassen {Ben Hassine} and Thibaud Barreau and Peter Tankov},
  title={Climate Finance Bench: A Benchmark for {RAG} over Corporate Climate Disclosures},
  journal={arXiv preprint arXiv:2505.22752},
  year={2025}
}

@inproceedings{lima2025rag,
  author={Rafael {Teixeira de Lima} and Shubham Gupta and Cesar {Berrospi Ramis} and Lokesh Mishra and Michele Dolfi and Peter Staar and Panagiotis Vagenas},
  title={Dataset Taxonomy and Generation Strategies for Evaluating {RAG} Systems},
  booktitle={Proceedings of the 31st International Conference on Computational Linguistics: Industry Track},
  year={2025},
  url={https://aclanthology.org/2025.coling-industry.4}
}

@article{esglens2026,
  author={Tsung-Yu Yang and Meng-Chi Chen},
  title={{ESGLens}: An {LLM}-Based {RAG} Framework for Interactive {ESG} Report Analysis and Score Prediction},
  journal={arXiv preprint arXiv:2604.19779},
  year={2026},
  url={https://arxiv.org/abs/2604.19779}
}

@inproceedings{ares2024,
  author={Jon Saad-Falcon and Omar Khattab and Christopher Potts and Matei Zaharia},
  title={{ARES}: An Automated Evaluation Framework for Retrieval-Augmented Generation Systems},
  booktitle={Proceedings of the 2024 Conference of the North American Chapter of the Association for Computational Linguistics: Human Language Technologies},
  pages={338--354},
  year={2024},
  doi={10.18653/v1/2024.naacl-long.20}
}

@inproceedings{liu2023geval,
  author={Yang Liu and Dan Iter and Yichong Xu and Shuohang Wang and Ruochen Xu and Chenguang Zhu},
  title={G-Eval: {NLG} Evaluation Using {GPT}-4 with Better Human Alignment},
  booktitle={Proceedings of the 2023 Conference on Empirical Methods in Natural Language Processing},
  pages={2511--2522},
  year={2023},
  doi={10.18653/v1/2023.emnlp-main.153}
}

@inproceedings{zhu2025judgelm,
  author={Chi Zhu and Yulong Chen and Xiaoyu Shen and Miao Xu and Qian Zhao and Jia Wei and Jianshu Huang and others},
  title={JudgeLM: Fine-Tuned Large Language Models Are Scalable Judges},
  booktitle={International Conference on Learning Representations},
  year={2025},
  url={https://openreview.net/forum?id=xsELpEPn4A}
}

@article{kim2024prometheus,
  author={Seungone Kim and Dongha Lee and Suhwan Kim and Hwiyeol Park and Minbyul Jeong and Seunghyun Yoon and others},
  title={Prometheus: Inducing Fine-Grained Evaluation Capability in Language Models},
  journal={arXiv preprint arXiv:2310.08491},
  year={2024}
}

@article{gu2024llmjudge,
  author={Yuxian Gu and Xiaochuang Han and Zheng Li and others},
  title={A Survey on {LLM}-as-a-Judge},
  journal={arXiv preprint arXiv:2411.15594},
  year={2024}
}

@misc{deepeval_docs,
  author={{Confident AI}},
  title={{DeepEval} Documentation: {RAG} Evaluation},
  howpublished={\url{https://deepeval.com/docs/getting-started-rag}},
  year={2026},
  note={Accessed: 2026-05-18}
}

@misc{ESRS2023,
  author={{European Commission}},
  title={Commission Delegated Regulation (EU) 2023/2772 supplementing Directive 2013/34/EU as regards sustainability reporting standards ({ESRS})},
  year={2023},
  howpublished={Official Journal of the European Union, L series},
  note={European Sustainability Reporting Standards, developed by {EFRAG}}
}

\end{document}